\documentclass{article}

    \PassOptionsToPackage{numbers, compress}{natbib}
    \usepackage[final]{neurips_2020}

\usepackage[utf8]{inputenc} % allow utf-8 input
\usepackage[T1]{fontenc}    % use 8-bit T1 fonts
\usepackage{url}            % simple URL typesetting
\usepackage{booktabs}       % professional-quality tables
\usepackage{amsfonts}       % blackboard math symbols
\usepackage{nicefrac}       % compact symbols for 1/2, etc.
\usepackage{microtype}      % microtypography

\usepackage{amsmath,amsfonts,bm}

\def\eqref#1{equation~\ref{#1}}
\def\1{\bm{1}}

\DeclareMathAlphabet{\mathsfit}{\encodingdefault}{\sfdefault}{m}{sl}
\SetMathAlphabet{\mathsfit}{bold}{\encodingdefault}{\sfdefault}{bx}{n}

\usepackage{algorithm}
\usepackage[noend]{algpseudocode}
\usepackage{amsmath}
\usepackage{amssymb}
\usepackage{amsthm}
\usepackage{courier}  % DO NOT CHANGE THIS
\usepackage{dsfont}
\usepackage{enumitem}
\usepackage{graphicx} % DO NOT CHANGE THIS
\usepackage{helvet} % DO NOT CHANGE THIS
\usepackage[colorlinks = true,
            linkcolor = cyan,
            urlcolor  = cyan,
            citecolor = blue,
            anchorcolor = blue]{hyperref}
\usepackage{listings}
\usepackage{mathtools}
\usepackage{multirow}
\usepackage{perpage} %the perpage package
\usepackage{pifont}% http://ctan.org/pkg/pifont
\usepackage{subfig}

\usepackage{times}  % DO NOT CHANGE THIS
\usepackage[normalem]{ulem}
\usepackage{wrapfig}
\usepackage{xcolor}
\usepackage{xspace}

\newcommand{\originalgrumbler}[2]{\begin{quote}\textcolor{blue}{\sl{\bf #1 says:} #2}\end{quote}}
\newcommand{\grumbler}[2]{\originalgrumbler{#1}{#2}}
\newcommand{\minjia}[1]{\grumbler{Minjia}{#1}}

\newcommand\later[1]{\begin{quote}\textcolor{green}{\textbackslash \textbf{later\{}} #1 \textcolor{green}{\}}\end{quote}}
\newcommand\notes[1]{\begin{quote}\textcolor{green}{\textbackslash \textbf{notes\{}} #1 \textcolor{green}{\}}\end{quote}}

\newcommand{\sfsmaller}{}

\newtheorem{defn}{Definition}[section]

\newcommand{\gatedtransformer}{Switchable-Transformer\xspace}

\newcommand{\gt}{ST\xspace}

\newcommand{\pst}{PLD\xspace}

\newcommand{\curriculumdrop}{progressive layer dropping\xspace}

\newcommand{\curriculum}{progressive\xspace}
\newcommand{\Curriculum}{Progressive\xspace}

\newcommand{\progressiveschedule}{progressive schedule\xspace}

\newcommand{\LayerDrop}{Layer\_Dropping}
\newcommand{\layerdrop}{layer dropping\xspace}

\renewcommand{\grumbler}[2]{}
\renewcommand{\notes}[1]{}
\renewcommand{\later}[1]{}
\renewcommand{\href}[2]{}

\newcommand{\outline}[1]{\grumbler{outline}{#1}}
\renewcommand{\outline}[1]{}

\usepackage[skip=1pt]{caption}
\usepackage{setspace}
\usepackage{enumitem}

\usepackage{titlesec}

\newcommand{\setvspace}[2]{%
  #1 = #2
  \advance #1 by -1\parskip}

\makeatletter % Control space around theorem and lemma
\def\thm@space@setup{%
  \thm@preskip=3pt
  \thm@postskip=\thm@preskip % or whatever, if you don't want them to be equal
}
\makeatother

\makeatletter
\g@addto@macro\normalsize{%
  \setlength\abovedisplayskip{1pt}
  \setlength\belowdisplayskip{1pt}
  \setlength\abovedisplayshortskip{1pt}
  \setlength\belowdisplayshortskip{1pt}
}
\makeatother



\title{Accelerating Training of Transformer-Based Language Models with Progressive Layer Dropping}
\author{%
    Minjia Zhang\qquad Yuxiong He \\
  Microsoft Corporation\\  
  \texttt{\{minjiaz,yuxhe\}@microsoft.com} \\
}

\begin{document}

\maketitle

\begin{abstract}

Recently, Transformer-based language models have demonstrated remarkable performance across many NLP domains. 
% many downstream natural language processing tasks. 
% pre-trained with proper unsupervised prediction tasks offer excellent results 
However, the unsupervised pre-training step of these models suffers from unbearable overall computational expenses.
% becoming very resource and time consuming. 
Current methods for accelerating the pre-training either rely on massive parallelism with advanced hardware or are not applicable to language modeling.
In this work, we propose a method based on \emph{progressive layer dropping} that speeds the training of Transformer-based language models, not at the cost of excessive hardware resources but from model architecture change and training technique boosted efficiency. 
% which can be leveraged to train the bidirectional Transformer (BERT) model in a more efficient way. In particular, we introduce an architectural change -- \gatedtransformer block, which allows BERT to be more robustly trained with layer removal. Meanwhile, we propose to use a progressive layer drop (\pst) learning strategy that continuously reduces the computation cost during training.
Extensive experiments on BERT show that the proposed method achieves 
% a 25\% reduction of computation cost in FLOPS and 
a 24\% time reduction on average per sample and allows the pre-training to be 2.5$\times$ faster than the baseline to get a similar accuracy on downstream tasks. While being faster, our pre-trained models are equipped with strong knowledge transferability, achieving comparable and sometimes higher GLUE score than the baseline when pre-trained with the same number of samples.

% Recently, Transformer-based language models have demonstrated remarkable performance across many NLP domains. 
% However, the unsupervised pre-training step of these models suffers from unbearable overall computational expenses.
% Current methods for accelerating the pre-training either rely on massive parallelism with advanced hardware or are not applicable to language modeling.
% In this work, we propose a method based on \emph{progressive layer dropping} that speeds the training of Transformer-based language models, not at the cost of excessive hardware resources but from model architecture change and training technique boosted efficiency. 
% Extensive experiments on BERT show that the proposed method achieves 
% a 24\% time reduction on average per sample and allows the pre-training to be 2.5X faster than the baseline to get a similar accuracy on downstream tasks. While being faster, our pre-trained models are equipped with strong knowledge transferability, achieving a 1.1 point higher GLUE score than the baseline when pre-trained with the same number of samples.

\minjia{it retains more than 99\% accuracy on SQuAD 2.0 and several GLUE benchmark tasks using 50\% of the Transformer parameters and computations of the teacher model.Give an averaged GLUE store and then measure the accuracy retain rate. -- check MiniLM}

% Recently, Transformer-based language models have demonstrated remarkable performance across many NLP domains. However, the unsupervised pre-training step of these models suffers from unbearable overall computational expenses.
% Current methods for accelerating the pre-training either rely on massive parallelism with advanced hardware or are not applicable to language modeling.
% In this work, we propose a method based on \emph{progressive layer dropping} that speeds the training of Transformer-based language models, not at the cost of excessive hardware resources but from model architecture change and training technique boosted efficiency. Extensive experiments on BERT show that the proposed method achieves a 24\% time reduction on average per sample and allows the pre-training to be 2.5$\times$ faster than the baseline to get a similar accuracy on downstream tasks. While being faster, our pre-trained models are equipped with strong knowledge transferability, achieving comparable and sometimes higher GLUE score than the baseline when pre-trained with the same number of samples.

\end{abstract}

\section{Introduction}
\label{sec:intro}

% \minjia{Story: First demonstrate change of layer-norm from Hessian fine-scale perspective. Then training deep model, then show stochastic depth helps. Similar to ALBERT, which share weights. Stochastic depth bert should be focusing on improved training speed. I can either show BERT-base L12 have better performance than BERT-large L24, or I can show BERT-base L24 or L36 has better performance than BERT-large.}

% \minjia{Separate two-phases: First phase, full network, second phase: stochastic drops. This is just a special case of drop schedule. In other words, explore the temporal schedule beyond the depth and model architecture itself. \minjia{Addressed through extending the work to "curriculum layerdrop"}}

% However, this is difficult because for a given number of parameters, it requires a lot of domain knowledge to make it efficient. 

% Recent progress in natural language processing has been fueled by increasing complex and large language models and transfer learning. 
Natural language processing (NLP) tasks, such as natural language inference~\cite{xlnet,roberta} and question answering~\cite{bert,commonsenseqa,bertserini}, have achieved great success with the development of neural networks. It has been demonstrated recently that Transformer-based networks have obtained superior performance in many NLP tasks (e.g., the GLUE benchmark~\cite{glue} and the challenging multi-hop reasoning task~\cite{multi-hop-reasoning}) than recurrent neural networks or convolutional neural networks. BERT trains a deep bidirectional Transformer and obtains outstanding results with transfer learning~\cite{bert}. RoBERTa~\cite{roberta}, which is a robustly optimized version of BERT trained with more steps and larger corpora, achieves state-of-the-art results on 9 GLUE tasks.  Megatron-LM~\cite{megatron-lm} further advances the state-of-the-art in NLP by significantly increasing the size of BERT model. Finally, there are multiple research proposing different enhanced versions of Transformer-based networks, such as GPT-2/3~\cite{gpt-2,gpt-3}, XLNet~\cite{xlnet}, SpanBERT~\cite{span-bert},  BioBERT~\cite{biobert}, UniLM~\cite{unilm}, Turing-NLG~\cite{turing-nlg}, and T5~\cite{T5}. 
% first leverages masked language modeling to learn universal language representations through unsupervised pre-training on massive web data, followed by a subsequent task-specific supervised fine-tuning, which turns to be more sample efficient and often significantly boosts performance of downstream tasks, such question and answering, text similarity, textual entailment, and sentiment analysis~\cite{bert}.
% More importantly, pretrained models have not yet reached its upper bound. 
%  through more training steps and larger corpora~\cite{roberta,scaleing-law-nlp}, 
% and its extensions such as XLNet~\cite{xlnet}, RoBERTa~\cite{roberta},  SpanBERT~\cite{span-bert}, and UniLM~\cite{unilm}
% Moreover, most of the current approaches can be further improved by more training steps and larger corpora~\cite{roberta,scaleing-law-nlp}, and the state-of-the-art in NLP can be further advanced by increasing the size of models, such as Megatron-LM~\cite{megatron-lm} and Turing-NLG~\cite{turing-nlg}. 
Due to the exciting prospect, pre-training Transformer networks with a large corpus of text followed by fine-tuning on specific tasks has become a new paradigm for natural language processing.
% has become a problem of great interest on how to advance these pre-trained language models. 
% A recent work of Qiu et. al. has an excellent survey on the state-of-the-art~\cite{}. 
% Pre-training language models then fine-tuning on downstream tasks has become a new paradigm for natural language processing (NLP). 

% Furthermore, recent evidence from these improvements reveals that a large network is of crucial importance for achieving state-of-the-art performance and in general there are increasing gains in predictive performance as model size and datasets grow~\cite{gpt-2,roberta,megatron-lm}. 
% Nevertheless, human effort has been shifted to pre-train larger models to have better NLP models.  

% where the last layers of the model are removed/retrained, which has already lead to a series of breakthroughs in language representation learning due to its capability 
% doing parallel recurrence and transfer learning to a variety of NLP tasks. The pretrained language models also can effectively model textual variations and distributional similarity. Therefore, they can make subsequent task-speciﬁc training more sample efficient and often significantly boost performance in downstream tasks.

Despite great success, a big challenge of Transformer networks comes from the training efficiency -- even with self-attention and parallelizable recurrence~\cite{transformer}, and extremely high performance hardware~\cite{tpu}, the pre-training step still takes a significant amount of time. 
% This is because the pre-training is done with an unprecedentedly massive amount of data and model parameters. Since no human generated labels are required, the size of the available training data easily scales
% up to billions of tokens. 
% And due to the large volume of data, large model size is also required to capture textual variations and distributional similarity. 
% Although the original BERT uses Wikipedia and Bookcorpus data for pre-training~\cite{bert}, many 
% Recent works also find that the larger dataset used for pre-training, the larger gains downstream tasks will benefit from~\cite{roberta,scaleing-law-nlp}, but that also takes a prohibitively large amount of time and computational resources.
% raise  
% So the amount of time required to train larger models is going to be even more. 
% As the larger dataset is considered, the process of pre-training brings 
% challenges in terms of both of the budget (resources) and time for training such models. 
% As a result, even on high performance hardware, it can take 96 hours on 64 TPUv2 chips~\cite{bert} or 1024 V100 GPU 1 day to pre-train BERT~\cite{roberta}. 
% On the other hand, training these huge models is a challenging problem, where 
To address this challenge, mixed-precision training is explored~\cite{megatron-lm,mixed-precision-training}, where the forward pass and backward pass are computed in half-precision and parameter update is in single precision. However, it requires Tensor Cores~\cite{tensorcore}, which do not exist in all hardware. Some work resort to distributed training~\cite{gpipe,mesh-tensorflow,megatron-lm}. However, distributed training uses large mini-batch sizes to increase the parallelism, where the training often converges to sharp local minima with poor generalizability even with significant hyperparameter tuning~\cite{on-large-batch-training}. Subsequently, Yang et al. propose a layer-wise adaptive large batch optimizer called LAMB~\cite{lamb}, allowing to train BERT with 32K batch size on 1024 TPU chips. However, this type of approach often requires dedicated clusters with hundreds or even thousands of GPUs and sophisticated system techniques at managing and tuning distributed training, not to mention that the amount of computational resources is intractable for most research labs or individual practitioners.

In this paper, we speedup pre-training Transformer networks by 
% introducing an architectural change and improving the network's learning procedure with a limited computational resource. Therefore,
% a more practical direction is to 
exploring architectural change and training techniques, not at the cost of excessive hardware resources.
Given that the training cost grows linearly with the number of Transformer layers, one straightforward idea to reduce the computation cost is to reduce the depth of the Transformer networks. However, this is restrictive as it often results in lower accuracy in downstream tasks compared to full model pre-training, presumably because of having smaller model capacities~\cite{distill-bert,tiny-bert}. Techniques such as Stochastic Depth have been demonstrated to be useful in accelerating supervised training in the image recognition domain~\cite{stochastic-depth}. However, we observe that stochastically removing Transformer layers destabilizes the performance and
% On the other hand, prior studies~\cite{dropout,stochastic-depth} suggest that it is not
% necessary to have every unit in the network participate in the training process at every training step. However, training Transformer networks such as BERT with layer removal requires extra effort. In particular, pre-training BERT can already
easily results in severe consequences such as model divergence or convergence to bad/suspicious local optima. Why are Transformer networks difficult to train with stochastic depth? Moreover, can we speed up the (unsupervised) pre-training of Transformer networks without hurting downstream performance?
To address the above challenges,
we propose to accelerate pre-training of Transformer networks by 
% with a stochastic training method (\pst). 
% that reduces the training time while achieving similar performance to models trained from scratch.
% improves the convergence rate of the BERT pre-training step.
% In particular, we make 
making the following contributions. (i) We conduct a comprehensive analysis to answer the
question: what makes Transformer networks difficult to train with stochastic depth. We find that both the choice of Transformer architecture as well as training dynamics would have a big impact on layer dropping. (ii) We propose a new architecture unit, called the \emph{\gatedtransformer} (\gt) block, that
% which performs \emph{identity mapping reordering} and contains a \emph{gate function} that controls whether to switch on/off an entire \gt block. \gt blocks 
not only allows switching on/off a Transformer layer for only a set portion of the training schedule, excluding them from both forward and backward pass but also stabilizes Transformer network training. (iii) We 
% We replace the original transformer block with the \gt block and train from scratch. 
% During training we then use 
further propose a \emph{\progressiveschedule} to add extra-stableness for pre-training Transformer networks with layer dropping -- 
% a schedule that determines the fraction of \gt blocks to bypass for each mini-batch. 
our schedule smoothly increases the layer dropping rate for each mini-batch as training evolves by adapting in time the parameter of the Bernoulli distribution used for sampling. Within each gradient update, we distribute a global \layerdrop rate across all the \gt blocks to favor different layers. (iv) We use BERT as an example, and we conduct extensive experiments 
to show that
% verify the efficiency and effectiveness of our proposed method.
% Specifically, our 
% According to our results, we find that during pre-training, our
the proposed method not only allows to train BERT 24\% faster 
% with reduced depth in a more stable way but it also requires 25\% less training computation in terms of FLOPS and is XX\%
% faster 
than the baseline under the same number of samples but also allows the pre-training to be 2.5$\times$ faster to get similar accuracy on downstream tasks. 
% to get higher accuracy, and achieves 2$\sim$3 times speedup with comparable accuracy on BookCorpus and Wikipedia data.
% the same validation
% accuracy. Second, 
Furthermore, we evaluate the generalizability of models pre-trained with the same number of samples as the baseline, and we observe  
% our final model is competitive and even
% better than the baseline model on several downstream tasks.
% We observe 
that while faster to train,
% less computation is used in pre-training, 
% the generalizability is not affected. Particularly, our approach is more robust on the choice of learning rates for both pre-training and fine-tuning, 
% and
our approach achieves a 1.1\% higher GLUE score than the baseline, indicating a strong knowledge transferability. 
% we either outperform or are competitive (on average 1.1pt higher) with the baseline models. 
% on various downstream tasks

% smoothly increases training difficulties by taking into account both temporal and architectural characteristics of BERT training, which preserves model accuracy. 
% and it smoothly increases training difficulties over training steps and over different layers 
% This approach is possible because, as we observe, BERT training is a dynamic process, both across layers and during the entire period of training. Our approach takes into account both temporal and architectural characteristics of BERT training and smoothly increases training difficulties over training steps and over different layers. 
% As a result, the method reduces the computation cost as it shorten the expected depth of BERT significantly during training. 

% The effect is an extended BERT model with a small expected depth during training. 

% \minjia{One thing that is unclear is what specific about Bert that motivates the our design. Without it, it feels like we are just applying stochastic depth. The special part about BERT: (1) warm-up, (2) self-attention. (3) after identity reordering, some layers are not learning.\minjia{Addressed by adding the BERT pre-training analysis.}}

\later{Our \progressiveschedule involves two aspects. From the temporal
aspect, we observe that BERT training has two noticeable phases -- a warmup phase in the beginning, and a cool-down phase that employs a learning rate decay schedule. Training, in the beginning, is more difficult than the later phase because there is out-of-bound variance, as also observed by some recent work~\cite{radam}, and the gradient variance becomes more stable after the warmup phase. 
Based on this observation, we dynamically increase the number of
\gt blocks that are dropped as a function of the number of gradient updates to adapt to the dynamic process of training. Since dropping \gt blocks at early training steps introduces additional perturbation and complexities that make training difficult, we smoothly increase the dropping rate as training evolves by adapting in time the parameter of the Bernoulli distribution used for sampling. Within each gradient update, we distribute a global layer-drop rate across all the \gt blocks to favor lower layer training. This is because we observe that lower layers are more subject to vanishing gradients~\cite{transformer-xl}, especially as the training proceeds, and they are hard to learn because of insufficient weight changes. By stochastically dropping upper layers more often, loss signals are directly propagating to lower layers, allowing lower layers to learn faster. \minjia{There are some issues in this text. We now know that unbalanced gradients only exist in the beginning of the training, so "especially as the training proceeds" part is not true.}}

\notes{Designing optimizer is like optimizing the search algorithm in ANN graph. Although it may lead to improvement, the underlying model structure (the graph property) often brings improvement that is more fundamental.}

\later{
More interestingly, we find that our proposed method consistently gains improved performance and accuracy as we scale up the depth of the BERT model, indicating the robustness and scalability of our approach.}

% Based on these changes, we present competitive results on \ResBert, which is much easier to train and generalizes better than the original BERT model. We empirically demonstrate that \ResBert outperforms the original BERT-large model, even under the restricted condition of maintaining computational complexity and model size over the Wiki and Bookcorpus datasets.
% using a 72-layer BERT, where our approach greatly improves generalizability and training efficiency, for which the counter part of ~\cite{bert} starts to overfit. 
% We emphasize that while it is relatively easy to increase accuracy by increasing capacity (going deeper or wider), our approach increases the accuracy while maintaining (or reducing) complexity. 

% Our results suggest that there is a large room to exploit the dimension of training acceleration for BERT like models, a key to its commercial success. 

\later{
In summary, our contributions in this paper are as follows:

\begin{itemize}
    \item We propose an efficient training method to accelerate BERT training using \gatedtransformer and \curriculum layer-drop, which saves training computation while retaining the performance. 
    % method to accelerate the pre-training process of BERT model by a careful designed spatial and temporal layer drop schedule.
    % \item We propose a novel architecture for BERT blocks that allows for residual connections with significantly improved performance.
    \item We validate the proposed method on BERT pre-training on large-scale dataset (Wikipedia + Bookcorpus).
    \item We verify that although less training computation is used, the generalization ability of the pre-trained model using our approach is not affected in a variety of downstream tasks, even with certain improvement in some experiments. 
    \item We show that our approach is scalable and consistently leads to improved performance for downstream tasks as the model depth increases.
    % conduct extensive experiments to show that the proposed approach is effective in accelerating BERT training, achieving comparable accuracy on down-stream tasks, and being possible to scale as the model depth increases.
\end{itemize}
}

% \paragraph{Contributions}

% Our contributions are as follows. 
% \begin{itemize}
% \item We show that depth is an important factor to acquire competitive NLM models with the transformer. 
% \item In order to facilitate training of very deep configurations, we propose a variation of stochastic depth for the Transformer inspired by the Stochastic Residual Network for image classification.
% \end{itemize}

% We discovered that its ability to regularize is the key contribution to obtain the state-of-the-art result for the wikipedia + bookcorpus dataset.

\section{Background and Related Work}
\label{sec:background}

% \subsection{Unsupervised Pre-training for NLP}

% Pre-trained word vectors~\cite{word2vec,glove} have been considered as a major component of most state-of-the-art NLP architectures, especially for those tasks where the amount of labeled data is not large enough~\cite{cnn-word2vec,tree-lstm-word2vec}. However, these learned word vectors only capture the semantics of individual words whereas the rich syntactic and semantic structures of phrases are not effectively exploited. 
% Pre-trained contextual representation overcomes the shortcomings of traditional word vectors by considering its surrounding context. Previous contextual embeddings such as ELMo are trained with stacked LSTMs~\cite{contextual-embedding}. However, LSTMs are processed sequentially, which severely limit the amount of parallelism that can be exploited. Transformer network uses advanced self-attention units instead of LSTMs in language modeling~\cite{transformer}. Devlin et al. further develops a masked language modeling task and achieves state-of-the-art performance on multiple natural language understanding tasks through transfer learning~\cite{bert}. As masked language modeling requires no human labeling effort, billions of sentences on the web can be used to train a very deep network. Therefore, a major challenge in learning such a model is training efficiency.  

\href{https://towardsdatascience.com/deep-learning-isnt-hard-anymore-26db0d4749d7}{Deep learning isn’t hard anymore}

\href{https://ruder.io/transfer-learning/
}{Transfer Learning - Machine Learning's Next Frontier}

\notes{Andrew Ng, chief scientist at Baidu and professor at Stanford, said during his widely popular NIPS 2016 tutorial that transfer learning will be -- after supervised learning -- the next driver of ML commercial success.}

% \subsection{Unsupervised Pre-training and Transformer Networks}

% Pre-trained word vectors have been considered as a major component of most state-of-the-art NLP architectures, especially for those tasks where the amount of labeled data is not large enough~\cite{word2vec,glove}. Prior work only capture the semantics of individual words~\cite{cnn-word2vec,tree-lstm-word2vec}. Some studies consist exploiting the syntactic and semantic structures of phrases by considering surrounding context using stacked LSTMs~\cite{contextual-embedding}, but LSTMs are processed sequentially, which severely limit the amount of parallelism that can be exploited. Transformer network overcomes this shortcoming through self-attention units instead of LSTMs in language modeling~\cite{transformer}.  

% Pre-training has always been an effective strategy to learn the parameters of deep neural networks, which are then fine-tuned on downstream tasks. 
Pre-training with Transformer-based architectures like BERT~\cite{bert} has been demonstrated as an effective strategy for language representation learning~\cite{roberta,xlnet,albert,megatron-lm}. The approach provides a better model initialization for downstream tasks by training on large-scale unlabeled corpora, which often leads to a better generalization performance on the target task through fine-tuning on small data.
% on downstream tasks. 
% recently become popular for language representation learning~\cite{roberta,xlnet,albert,megatron-lm}.
% It is effective because it provides a better model initialization that often leads to a better generalization performance on the target task and can be regarded as a kind of regularization to avoid overfitting on small data~\cite{why-pre-training-helps}. 
% Transformer 
% The breakthrough of pre-training for NLP came with Devlin et al., who developed a masked language modeling task called BERT and achieved state-of-the-art performance on multiple natural language understanding tasks~\cite{bert}, followed by various extensions and applications. A recent work of Qiu et. al. has an excellent survey on the pre-trained models for NLP~\cite{pre-training-survey}.
% We succinctly review the BERT model architecture and refer the reader to \cite{bert} for additional details. 
Consider BERT, which consists a stack of $L$ Transformer layers~\cite{transformer}.
Each Transformer layer encodes the the input of the i-th Transformer layer $x_i$ with $h_i = f_{LN}(x_i + f_{S-ATTN}(x_i))$, which is a multi-head self-attention sub-layer $f_{ATTN}$, and then by $x_{i+1} = f_{LN}(h_i + f_{FFN}(h_i))$, which is a feed-forward network $f_{FFN}$, where $x_{i+1}$ is the output of the i-th Transformer layer. Both sub-layers have an AddNorm operation that consists a residual connection~\cite{resnet} and a layer normalization ($f_{LN}$)~\cite{layer-norm}. The BERT model recursively applies the transformer block to the input to get the output.

\href{https://arxiv.org/pdf/1711.08393.pdf}{BlockDrop: Dynamic Inference Paths in Residual Networks}

\notes{Transformers essentially uses Convolutional Neural Networks together with attention models.}

\href{https://openreview.net/pdf?id=ByxRM0Ntvr}{ARE TRANSFORMERS UNIVERSAL APPROXIMATORS
OF SEQUENCE-TO-SEQUENCE FUNCTIONS?}

\href{https://towardsdatascience.com/transformers-141e32e69591}{How Transformers Work}

\notes{What are the “query”, “key”, and “value” vectors?
They’re abstractions that are useful for calculating and thinking about attention}

\notes{The second step in calculating self-attention is to calculate a score. Say we’re calculating the self-attention for the first word in this example, “Thinking”. We need to score each word of the input sentence against this word. The score determines how much focus/importance to place on other parts of the input sentence as we encode a word at a certain position. (MZ: Embeddings are therefore a function of its neighbors/context.) The fifth step is to multiply each value vector by the softmax score (in preparation to sum them up). The intuition here is to keep intact the values of the word(s) we want to focus on, and drown-out irrelevant words (by multiplying them by tiny numbers like 0.001, for example). Transformers basically work like that. There are a few other details that make them work better. For example, instead of only paying attention to each other in one dimension, Transformers use the concept of Multihead attention.}

\href{https://arxiv.org/pdf/1610.03022.pdf
}{Very deep convolutional networks for end-to-end speech recognition}

\href{https://arxiv.org/pdf/1909.11942.pdf
}{ALBERT: A LITE BERT FOR SELF-SUPERVISED LEARNING OF LANGUAGE REPRESENTATIONS}

% \section{Motivation and Challenges}

While the Transformer-based architecture has achieved breakthrough results in modeling sequences for unsupervised language modeling~\cite{bert,gpt-2}, previous work has also highlighted the training difficulties and excessively long training time~\cite{roberta}. 
% problematic, or even counterproductive.
To speed up the pre-training, ELECTRA~\cite{electra} explores the adversarial training scheme by replacing masked tokens with alternatives sampled from a generator framework and training a discriminator to predict the replaced token. This increases the relative per-step cost, but leads to fewer steps, leading to the overall reduced costs. 
% whether a replaced token is from a generator, 
% whereas we speed up the pre-training speed through an architectural change and a schedule for layer dropping. Working together, they could reduce the end-to-end pre-training time even further. 
Another line of work focus on reducing the per-step cost. 
Since the total number of floating-point operations (FLOPS) of the forward and backward passes in the BERT pre-training process is linearly proportional to the depth of the Transformer blocks, reducing the number of Transformer layers brings opportunities to significantly speed up BERT pre-training. To show this, we plot the FLOPS per training iteration in Fig.~\ref{fig:expected-GFLOPS-curve-graph}, assuming we can remove a fraction of layers at each step. Each line in the figure shows the FLOPS using different layer removal schedules. Regardless of which schedule to choose, the majority of FLOPS are reduced in the later steps, with the rate of keep probability saturating to a fixed value $\bar{\theta}$ (e.g., 0.5). We will describe our schedule in Section~\ref{subsec:curriculum-schedule}.

\begin{figure}[h!]
 \centering
 \begin{minipage}[c]{0.32\textwidth}
    \includegraphics[scale=0.38, keepaspectratio=true]{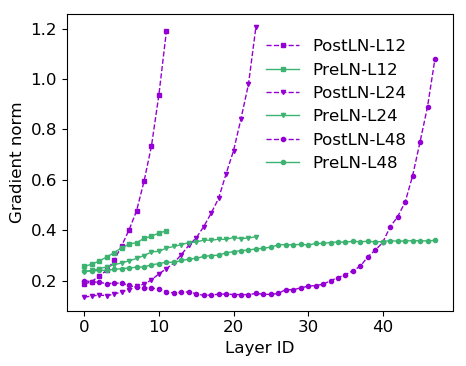}  
    \caption{The norm of the gradient with respect to the weights, with PostLN and PreLN.}
    \label{fig:stability-gradient-norm}
 \end{minipage}%
 \hfill
  \begin{minipage}[c]{0.32\textwidth}
    \includegraphics[scale=0.38, keepaspectratio=true]{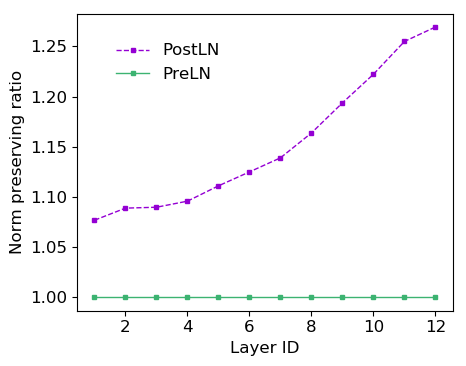}
    \caption{The norm preserving ratio with respect to the inputs, with PostLN and PreLN.}
    \label{fig:grad-norm-preserving}
 \end{minipage}%
 \hfill
 \begin{minipage}[c]{0.32\textwidth}
    \includegraphics[scale=0.38, keepaspectratio=true]{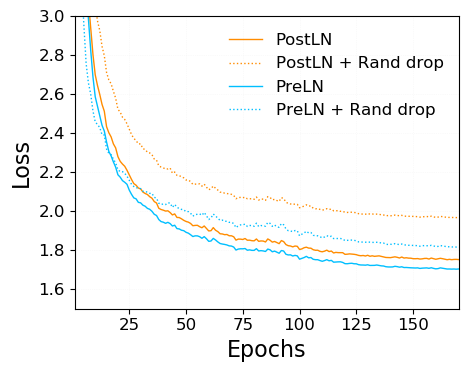}
    \caption{Lesioning analysis with PostLN and PreLN.}
    \label{fig:leisioning-analysis}
\end{minipage}
\end{figure}

% \minjia{Motivation for progressive layer drop.}

%  However, BERT is too unstable to train with stochastic depth. 

Despite the FLOPS reduction, directly training models like BERT with a smaller depth incurs a significant loss in accuracy even with knowledge distillation~\cite{distill-bert,tiny-bert}. Prior work~\cite{progressive-stacking} proposes to accelerate pre-training by first training a 3-layer BERT model and then growing the network depth to 6-layer and subsequently 12-layer. 
However, the number of steps required at each depth before the network growth is not known a prior, making applying this approach challenging in practice. On the other hand, stochastic depth has been successfully demonstrated to train deep models with reduced expected depth~\cite{stochastic-depth,reduce-depth-on-demand}. However, we observe that directly pre-training BERT with randomly dropping $f_{ATTN}$ and $f_{FFN}$ converges to bad/suspicious local optima under the same hyperparameter setting. When increasing the learning rate, the training often diverges even by tuning the warmup ratio.
% is less robust on the choice of hyperparameters. For example, in a parameter grid search (e.g., varying learning rates and warmup ratio), pre-training BERT with stochastic depth (i.e., drop $f_{ATTN}$ and $f_{FFN}$ stochastically) diverges in 6 out of 10 settings. 
% % Specifically, it diverges with larger learning rates. 
% When it converges, it converges to bad/suspicious local optima. 
What causes the instability of BERT pre-training with layer drop?
% and how to avoid suboptimal performance in downstream tasks? 

% \minjia{Using MNLI/SST-2 as examples to show XX out of 15. The heat scores.}

%  A straightforward idea to accelerate BERT pre-training is to train the model with a reduced depth. However, training a BERT model with a reduced depth incurs significant loss in accuracy~\cite{distill-bert}.  On the other hand, stochastic depth has been demonstrated to train deep models with reduced expected depth~\cite{stochastic-depth}. 
 
% \minjia{Opportunity from stabilizing training }

\section{Preliminary Analysis}
\label{sec:analysis}

% \subsection{Unbalanced Gradients}

This section presents several studies that guided the design of the approach introduced in Section~\ref{sec:method}. We used BERT trained on Bookcorpus and Wikipedia dataset from Devlin et. al. with standard settings as the baseline~\footnote{Appendix~\ref{sec:hyperparameters} provides detailed training hyperparameters.}. First, we carry out a comparison between BERT with PostLN and PreLN. Our goal is to measure how effective these two methods at stabilizing BERT training.
% identify why BERT with stochastic depth is difficult to train. 
Our second analysis considers measuring the dynamics of BERT pre-training, including both spatial and temporal dimensions. Finally, we analyze the effect of the removal of the Transformer layers. This leads us to identify appealing choices for our target operating points. 

\subsection{Training Stability: PostLN or PreLN?} 
\label{subsec:training-stability}

We consider two variants of BERT, namely the PostLN and PreLN. The default BERT employs PostLN, with layer normalization applied after the addition in Transformer blocks. The PreLN changes the placement of the location of $f_{LN}$ by placing it only on the input stream of the sublayers so that 
${h}_i = x_i + f_{S-ATTN}(f_{LN}(x_i))$ and then $x_{i+1} = h_i + f_{FFN}(f_{LN}(h_i))$, which is a modification described by several recent works to establish identity mapping for neural machine translation~\cite{deep-Transformer,on-layer-norm,adaptive-inputs,sparse-Transformers,Transformer-without-tears}. Fig.~\ref{fig:stability-gradient-norm} reports the norm of gradients with respect to weights in backward propagation for both methods, varying the depth $L$ (e.g., 12, 24, 48). The plot shows that while PostLN suffers from unbalanced gradients (e.g., vanishing gradients as the layer ID decreases), PreLN eliminates the unbalanced gradient problem (solid green lines) and the gradient norm stays almost same for any layer. Furthermore,  Fig.~\ref{fig:grad-norm-preserving} shows that for PreLN the gradients with respect to input $x_i$ have very
similar magnitudes (norm preserving ratio close to 1) at different layers, which is consistent with prior findings that a neural model should preserve the gradient norm between layers so as to have well-conditioning and faster convergence~\cite{understanding-difficulty-of-training-dnn,norm-preservation}. 
% and this leads to well-conditioning and faster convergence~\cite{understanding-difficulty-of-training-dnn}
Indeed, we find that PostLN is more sensitive to the choice of hyperparameters, and training often diverges with more aggressive learning rates (more results in Section~\ref{sec:eval}), whereas PreLN avoids vanishing gradients and leads to more stable optimization. We also provide preliminary theoretical results in Appendix~\ref{sec:preln-analysis} on why PreLN is beneficial.

\minjia{This is superficial. Need math to support. Also, having one result figure here is good. Having the similarity results here is too much for one design component. The L2/Cosine results can be in the eval for robustness analysis.}

\later{
This is important for our approach, since our approach reduces BERT depth, we want to increase the learning rate as a shallower and less complex model learns faster and suffers less from the effect of large variance of gradients in its deeper counterpart~\cite{}.
}

\minjia{Layer drop is type of pruning. Without stabilizing the training, it can be destructive.}

% \begin{figure}[!ht]
%     \centering
%     \includegraphics[scale=0.5]{figs/vanishing-gradient-v2}
%     \caption{(a) The gradient norms with and without \gt blocks}
%     \minjia{One experiment we could add: PreLN stableness. I'm guessing it is even more stable.}
%     \minjia{Measure again the gradient norm. This time, measure the aggregated norms for the entire layer and then get the mean.}
%     \minjia{Beyond this figure. It might also worthwhile showing the gradient distribution -- show the unexpected large variance.\minjia{Did this experiment, but I did not observe very large variance with warmup enabled.}}
%     \label{fig:stableness-analysis}
% \end{figure}

% Through identity mapping reordering, the output $X_{out}$ becomes a recursive summation of the outputs from all previous layers. Therefore, gradients at skip connection boundary now can be decomposed into two additive terms: a term that propagates information directly without concerning any weight layers, and another term that propagates through the weight layers. The first term ensures that the information is directly propagated back to any shallower unit.

% "Identity mappings are very important for healthy back-propagation"~\cite{Transformer-without-tears}

\href{https://arxiv.org/pdf/1706.02515.pdf}{Self-Normalizing Neural Networks}

% So when input values get multiplied by weight values, their activation remains on scale 1. What does it do? It helps in graceful optimization of neural networks. Since the hidden activation functions don’t saturate that fast. And thus does not give near zero gradients early on in learning.

\href{https://towardsdatascience.com/what-is-weight-initialization-in-neural-nets-and-why-it-matters-ec45398f99fa}{Weight initialization}

% Inspired by recent work on layer-normalization, we modify the residual connections to the following form:

% \begin{equation}
% \label{eqn:original-residual-connection}
%     R(x) = M \times F(LayerNorm(x)) + x
% \end{equation}

% \minjia{Need to understand what LayerNorm does. Some paper mentioned that LayerNorm helps to control the magnitude of the layer. }

\href{https://leimao.github.io/blog/Layer-Normalization/}{Layer Normalization Explained}

\href{https://stats.stackexchange.com/questions/304755/pros-and-cons-of-weight-normalization-vs-batch-normalization}{Pros and cons of weight normalization vs batch normalization}

\href{https://papers.nips.cc/paper/8689-understanding-and-improving-layer-normalization.pdf}{Understanding and Improving Layer Normalization}

% \begin{equation}
% \label{eqn:pre-layer-norm}
%     R(x) = M * F(LayerNorm(x)) * \frac{1}{1 - p_t} + x
% \end{equation}

% \minjia{To make it more Bert specific, we can give the full equation of the Transformer block. Or we can do an expansion of the Transformer block, so it becomes clear what we mean.}

% \minjia{TODO: Zhang and Sennrich, 2019: propose RMSNorm which normalizes by
% root mean square. It’s faster than LayerNorm, achieves comparable result.}

% \paragraph{Analysis.} 

% It helps to understand if we give the equation of the "identity mapping reordering", which for each layer $l$:

% \begin{equation}
% \begin{split}
%     X_{l+1} &= X_{l} + Attn(LayerNorm(X_{l})) + FFN(X_{l} + Attn(LayerNorm(X_{l})))) \\
% \end{split}
% \end{equation}

% \begin{equation}
% \label{eqn:analysis-1}
% \begin{split}
%     X_{l+1} &= Attn(X_l) + dropout(W_2\cdot gelu(W_1\cdot LayerNorm(Attn(X_l)))) \\
% \end{split}
% \end{equation}

% Let's denote $S(X) = dropout(W_2\cdot gelu(W_1\cdot LayerNorm(X)))$, and we will have:

% \begin{equation}
% \begin{split}
%     X_{l+2} &= X_{l+1} + F(LayerNorm(X_{l+1}), W_{l+1}) + S(X_{l+1} + F(LayerNorm(X_{l+1}), W_{l+1})) \\
% \end{split}
% \end{equation}

\minjia{Can we simply the second and third term? \minjia{Done}}

\subsection{Corroboration of Training Dynamics}
\label{subsec:training-dynamics}

\begin{figure}[ht!]
\centering
\small
\subfloat[]{\includegraphics[scale=0.29, keepaspectratio=true]{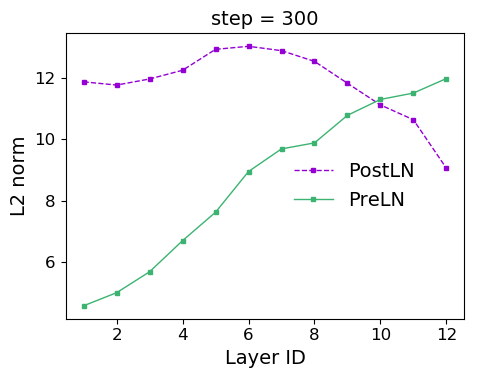}\label{fig:similarity-l2-norm-step300}} 
\subfloat[]{\includegraphics[scale=0.29, keepaspectratio=true]{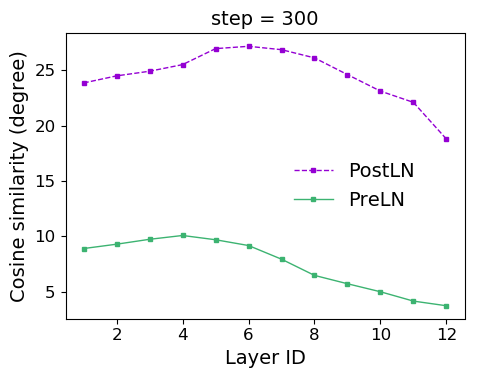}\label{fig:similarity-cosine-step300}} 
\subfloat[]{\includegraphics[scale=0.29, keepaspectratio=true]{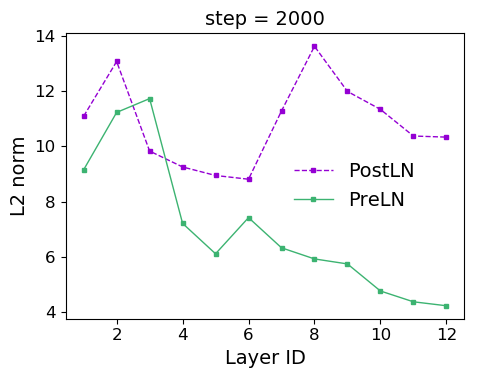}\label{fig:similarity-l2-norm-step2000}}
\subfloat[]{\includegraphics[scale=0.29, keepaspectratio=true]{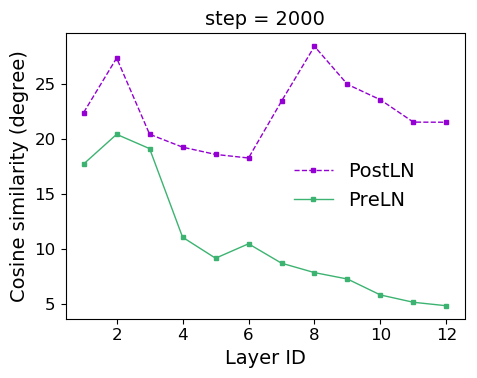}\label{fig:similarity-cosine-step2000}}
\caption{The L2 distance and cosine similarity of the input and output embeddings for BERT with PostLN and PreLN, at different layers and different steps. We plot the inverse of cosine similarity (arccosine) in degrees, so that for both L2 and arccosine, the lower the more similar.}
    \minjia{One experiment we could add: PreLN stableness. I'm guessing it is even more stable.}
    \minjia{Measure again the gradient norm. This time, measure the aggregated norms for the entire layer and then get the mean.}
\minjia{TODO: Also show that outputs are norm-preserving: the norm of the gradient with respect to the input is close to the norm of gradient with respect to the output.}
\label{fig:similarity-analysis}
\end{figure}

Hereafter we investigate the representation $x_i$ learned at different phases of BERT pre-training and at different layers. Fig.~\ref{fig:similarity-analysis} shows the L2 norm distances and cosine similarity, which measures the angle between two vectors and ignores their norms, between the input and output embeddings, with PostLN and PreLN, respectively. We draw several observations. 

First, the dissimilarity (Fig.~\ref{fig:similarity-l2-norm-step300} and Fig.~\ref{fig:similarity-cosine-step300}) stays high for both PostLN and PreLN at those higher layers in the beginning, and the L2 and cosine similarity seems to be less correlated (e.g., step = 300).  
% the variance of the similarity between input/output embeddings is still large for , at this stage is not very high for both with and without identity mapping reordering, especially for those higher layers, 
% For PreLN, the L1 dissimilarity increases over successive layers and become high at later layers. 
This is presumably because, at the beginning of the training, the model weights are randomly initialized, and the network is still actively adjusting weights to derive richer features from input data. Since the model is still positively self-organizing on the network parameters toward their optimal configuration, 
% Therefore, any co-adaption between layers should be preserved as positively representing the self-organization on the network parameters towards their optimal configuration. 
dropping layers at this stage is not an interesting strategy, because it can create inputs with large noise and disturb the positive co-adaption process.

% Furthermore, at the beginning of the training, the model weights are randomly initialized, and they are statistically independent and not co-adapted at all. Therefore, if any co-adaption between layers is displayed, it should be preserved as positively representing the self-organization on the network parameters towards their optimal configuration. In contrast, once the training has entered into the cooling down phase, the training process becomes more robust to layer drop. 

Second, as the training proceeds (Fig.~\ref{fig:similarity-l2-norm-step2000} and Fig.~\ref{fig:similarity-cosine-step2000}), although the dissimilarity remains relatively high and bumpy for PostLN, the similarity from PreLN starts to increase over successive layers, indicating that while PostLN 
% These results indicate that for PostLN, even in later phase of the training, the network 
is still trying to produce new representations that are very different across layers, the dissimilarity from PreLN is getting close to zero for upper layers, indicating that the upper layers are getting similar estimations. This can be viewed as doing an unrolled iterative refinement~\cite{iterative-estimation}, where a group of successive layers iteratively refine their estimates of the same representations instead of computing an entirely new representation. 
% In their studies, they report that deep ResNet can be viewed as a group of successive layers iteratively refine their estimates of the same representations instead of computing an entirely new representation. 
Although the viewpoint was originally proposed to explain ResNet, we demonstrate that it is also true for language modeling and Transformer-based networks. Appendix~\ref{sec:unrolled-analysis} provides additional analysis on how PreLN provides extra preservation of feature identity through unrolled iterative refinement. 

% Fig.~\ref{fig:similarity-l2-norm-step2000} and Fig.~\ref{fig:similarity-cosine-step2000} also show that the dissimilarity is getting close to zero for upper layers, indicating that the role of $f_{RT}$ is as iteratively refining a representation.  

% already starts to iteratively refine a good estimate (at least at those higher layers), leading to similar estimations that converge to a final representation. 

% compute new level of representations to provide a good estimate for the final representation, whereas higher layers

% At these layers, the network is probably learning low level representations that tend to be relatively simple and need little iterative refinement. Subsequent layers on the other likely need to handle more complex representations with numerous dependencies and therefore need more iterative refinement.
% which indicates that higher layers converge to similar estimation of representations. 

% iterative refinement of 
% similarity keeps increasing as moving to higher layers. 

% First, we observe that the transitions from layer to layer are much smoother for \gt blocks than for Transformer blocks. Second, the overall similarity of input/output embeddings is higher with \gt blocks than with Transformer blocks. Third, the similarity smoothly increases with identity mapping reordering after certain layers (e.g., after the third layer). 
\later{
These results indicate that for PostLN, even in later phase of the training, the network is still trying to produce new representations that are very different across layers. In contrast, for PreLN, the lower layers (e.g., 1 to 3) actively compute new level of representations to provide a good estimate for the final representation. At these layers, the network is probably learning low level representations that tend to be relatively simple and need little iterative refinement. Subsequent layers on the other likely need to handle more complex representations with numerous dependencies and therefore need more iterative refinement.}
% refine that estimate without changing the level of representation. 
% So if the first few layers have already learned simple contextual similarity information, then the rest of the layers will work at that level too. 

% with identity mapping reordering, the Transformer layers behave more like working together to estimate and iteratively refine a single level of representation after a given layer. The first layer
% provides a (rough) estimate for the representation, the second and the third layer learns new representations, and then subsequent layers refine the learned representation.

% Therefore, if the residual block ${f_{RT}}_i$ has a zero mean over the training set, then PreLN satisfies $\underset{x \in X}{\mathds{E}}[a_i - A_i] = 0$ and maintains identity mapping. 

\subsection{Effect of Lesioning}

% , indicating that it is valid to interpret the role of $f_{RT}$ as that of iteratively refining a representation.   

We randomly drop layers with a keep ratio $\theta = 0.5$ to test if dropping layers would break the training because dropping any layer changes the input distribution of all subsequent layers. The results are shown in Fig.~\ref{fig:leisioning-analysis}. As shown, removing layers in PostLN significantly reduces performance. Moreover, when increasing the learning rate, it results in diverged training.
In contrast, this is not the case for PreLN. Given that later layers in PreLN tend to refine an estimate of the representation, the model with PreLN has less dependence on the downsampling individual layers. As a result, removing Transformer layers with PreLN has a modest impact on performance (slightly worse validation loss at the same number of training samples). However, the change is much smaller than with PostLN. It further indicates that if we remove layers, especially those higher ones, it should have only a mild effect on the final result because doing so does not change the overall estimation the next layer receives, only its quality. The following layers can still perform mostly the same operation, even with some relatively little noisy input. Furthermore, as Fig.~\ref{fig:similarity-analysis} indicates, since the lower layers remain to have a relatively high dissimilarity (deriving new features), they should be less frequently dropped. Overall, these results show that, to some extent, the structure of a Transformer network with PreLN can be changed at runtime without significantly affecting performance. 

% Why is PreLN resilient to dropping layers but PostLN is not? 
% Given that later layers refine an already reasonable estimate of the representation, the model with PreLN has less dependence on the downsampling individual layers. 
% It indicates that removing layers, especially those higher ones, should have only a mild effect on the final result because doing so does not change the overall estimation the next layer receives, only its quality. The following layers can still perform mostly the same operation, even with some relatively little noisy input. Furthermore, as the lower layers are still actively deriving richer features from input data, they should be less frequently dropped.

% Training PreLN with layer drop improves resilience slightly (This is surprising as I thought it would decrease the resilience slightly); only the
% dependence on the downsampling layers seems to be reduced. By now, this is not surprising: we know that plain PreLN BERT already does not depend on individual layers.

\minjia{TODO: Add another section on how removing layers would have an impact to model training dynamics. Similar as Section 4.1 in https://arxiv.org/pdf/1605.06431.pdf.}

\section{Our Approach: Progressive Layer Dropping}
\label{sec:method}

% We explore both the depth and time dimension.

This section describes our approach, namely progressive layer dropping (\pst), to accelerate the pre-training of Transformer-based models.  We first present the Switchable-Transformer blocks, a new unit that allows us to train models like BERT with layer drop and improved stability. Then we introduce the progressive layer drop procedure.

\subsection{\gatedtransformer Blocks}

\begin{figure}[h!]
 \centering
 \begin{minipage}[c]{0.60\textwidth}
 %\centering
    \subfloat[Original]{\includegraphics[scale=0.5, keepaspectratio=true]{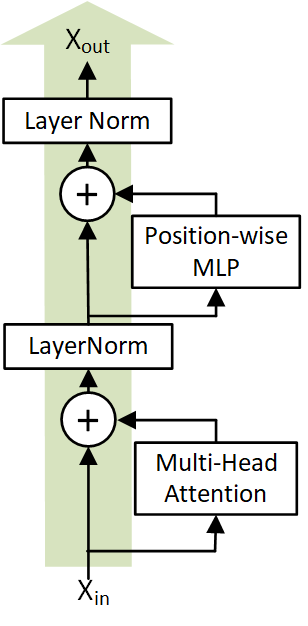}\label{fig:gated-transformer-v2a}} 
    \hspace{1.00em}
    \subfloat[Identity mapping reordering]{\includegraphics[scale=0.5, keepaspectratio=true]{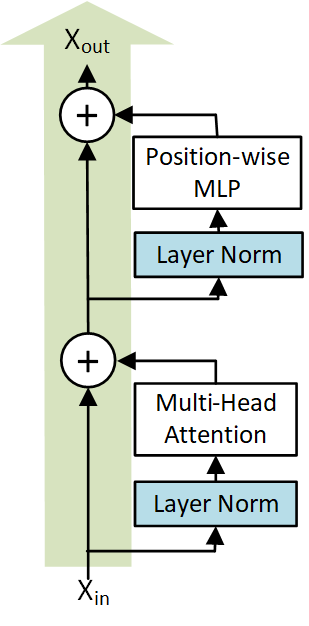}\label{fig:gated-transformer-v2b}} 
    \hspace{1.00em}
    \subfloat[Switchable Transformer]{\includegraphics[scale=0.5, keepaspectratio=true]{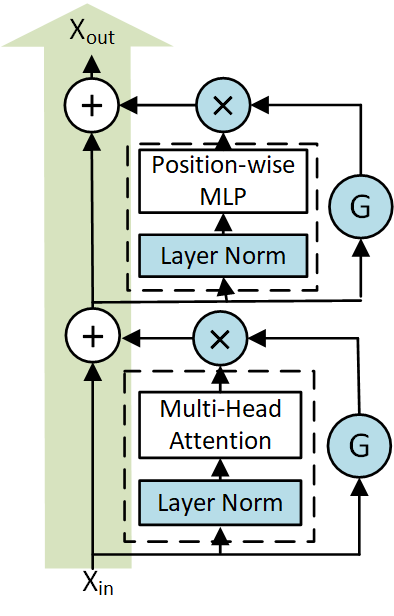}\label{fig:gated-transformer-v2c}} 
    \caption{Transformer variants, showing a single layer block.}\label{fig:gated-transformer-v2}
 \end{minipage}%
 \hspace{0.50em}
 \begin{minipage}[c]{0.34\textwidth}
 %\centering
    \includegraphics[scale=0.38, keepaspectratio=true]{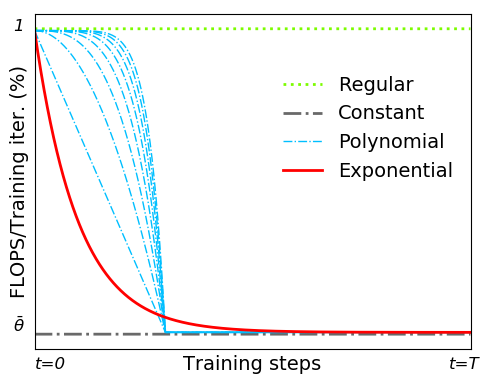}
    \caption{FLOPS per training iteration normalized to the baseline.}
    \label{fig:expected-GFLOPS-curve-graph}
\end{minipage}
\end{figure}

% Stochastic depth has been successful at reducing the execution time for image classification tasks~\cite{stochastic-depth}. Motivated by this, 
In this work, we propose a novel transformer unit, which we call "\gatedtransformer" (\gt) block. Compared with the original Transformer block (Fig.~\ref{fig:gated-transformer-v2a}), it contains two changes.

\paragraph{Identity mapping reordering.} The first change is to establish identity mapping within a transformer block by placing the layer normalization only on the input stream of the sublayers (i.e., use PreLN to replace PostLN) (Fig.~\ref{fig:gated-transformer-v2b}) for the stability reason described in Section~\ref{subsec:training-stability}. 
% a modification described in several previous works for image classification and neural machine translation~\cite{identity-mapping,deep-transformer,on-layer-norm,adaptive-inputs,sparse-transformers,transformer-without-tears}. 
% This change allows each layer to easily express the identity transformation.
% and creates an identity mapping from the input at the first layer to the output of the last layer, making backpropagation more efficient over depth. 
% In particular, through this change, the BERT model becomes a group of successive transformer layers iteratively refine their estimates
% of the same representations instead of computing an entirely new representation, making the training more stable.  

% \paragraph{Switchable gates.} 

% that replaces deterministic residual connections within the transformer block to stochastic ones, coupled with changes to the order of layer normalization in the sub-layers.
% % As demonstrated in, the "identity mapping reordering" and stochastic links are critical for making BERT training more efficient.  
% Fig.~\ref{fig:gated-transformer} depicts the structure of the \gt block, where we make the two changes.

\href{https://arxiv.org/pdf/1910.06764.pdf}{STABILIZING TRANSFORMERS
FOR REINFORCEMENT LEARNING}

\href{https://openreview.net/pdf?id=SylO2yStDr}{REDUCING TRANSFORMER DEPTH ON DEMAND WITH STRUCTURED DROPOUT}

% \subsubsection{Switchable Gates}

\paragraph{Switchable gates.} 

% We then replace deterministic residual connections within the transformer block to stochastic ones, coupled with changes to the order of layer normalization in the sub-layers.

% \minjia{1. Some blocks are redundant and are hard to learn. 2. We introduce stochastic links to address this problem. \minjia{Redundancy part is explained by unrolled iterative refinement.}}

% With the previous changes, the BERT model can be trained more robustly.
% suffers less from the vanishing gradient problem and can be favourably trained with many layers. 
% However, given that the BERT consists of multiple sub-layers taking different paths through shortcut connections during training, there are likely redundant layers, especially in the cooling down phase. 
Next, we extend the architecture to include a gate for each sub-layer (Fig.~\ref{fig:gated-transformer-v2c}), which controls whether a sub-layer is disabled or not during training. In particular, for each mini-batch, the two gates for the two sublayers decide whether to remove their corresponding transformation functions and only keep the identify mapping connection, which is equivalent to applying a conditional gate function $G$ to each sub-layer as follows:

\minjia{Our curriculum should be based on the learning rate and some other model properties. The fact that both warmup and learning rate decay indicate that there are some intrinsic dynamics in ths process. A function of learning rate and variance? Somehow subsume the current setting would be good. Generalize the current settings. Epsilon that is controlled by the variance of adaptive learning rate. Think about RAdam. Learning rate decay schedule. In this work, we did not talk about the optimizer.}
\begin{equation}
\begin{split}
    h_{i} &= x_{i} + G_{i} \times f_{S-ATTN}(f_{LN}(x_{i})) \times \frac{1}{p_i} \\
    x_{i+1} &= h_{i}+ G_{i} \times f_{FFN}(f_{LN}(h_{i})) \times \frac{1}{p_i} \\
\end{split}
\end{equation}
% , which reduces the effective length of BERT during training.

% \minjia{TODO No need to separate the spatial and temporal schedule. Instead, we should just give a single formula. Then we can give the breakdown of it. Not sure if this works. I think this from bottom up, but now it seems it might be better to describe it using a top-down approach. If that's the case, the model architecture change can be part of the spatial dropout. Sometimes, proposing an idea feels like making a dish or a dough. One makes a dough by mixing different gradients and knead it so that it becomes something delicious. Only after showing the cooked dish, we can give out the recipe and how one do it step by step.}

% What can we do while training our models, that will help them generalize even better. 
% The high density of residual connections is the reason why Transformer 

\notes{According to Xiaodong, we cannnot say deeper models are subject to overfitting in the context of unsupervised learning. It is hard to explain. Theoretically, if the model size is infinite, then sure it will overfit. However, under the context of unsupervised learning, the possible combinations (e.g., different ways of applying masks) can be combinatorial explosive, making it difficult to overfit. It would be better to describe the motivation of applying this technique from the perspective of accelerating the training process perspective.}

% (MZ: This motivates the work of stochastic depth training which forces all layers to learn something useful). The method resembles Dropout~\cite{dropout}, in which the key idea is that the layers are randomly dropped during training. 
% In particular, given the following form:

% \begin{equation}
% \begin{split}
%     X_{l+2} &= X_{l+1} + F(LayerNorm(X_{l+1}), W_{l+1}) + S(X_{l+1} + F(LayerNorm(X_{l+1}), W_{l+1})) \\
% \end{split}
% \end{equation}

\href{https://openreview.net/forum?id=ByxRM0Ntvr
}{Are Transformers universal approximators of sequence-to-sequence functions?}

% \minjia{Change this to per layer based, so that M becomes something like $M_l$.\minjia{Addressed}}

% In equation~\ref{eqn:original-residual-connection}, the inner function F is either self-attention or feed-forward layers. The layer normalization as in \cite{layer-norm} keeps the magnitude of the hidden layers from growing large.

In our design, the function $G_i$ only takes 0 or 1 as values, which is chosen randomly from a Bernoulli distribution (with two possible outcomes), $G_i \sim B(1, p_i)$,
% \begin{equation}
%     G_i \sim B(1, p_i)
% \end{equation}
where $p_i$ is the probability of choosing 1. Because the blocks are selected with probability $p_i$ during training and are always presented during inference, we re-calibrate the layers' output by a scaling factor of $\frac{1}{p_i}$ whenever they are selected. 
% is assumed to be p, similar to dropout. 
% \minjia{TODO: P above should be replaced with a curriculum function.}
% Specifically, we can multiply the neuron input x by m ∼ Bernoulli(Φ(x)), where Φ(x) = P(X ≤
% x), X ∼ N (0, 1) is the cumulative distribution function of the standard normal distribution.
% \href{}{GELU}
% This way we have one hyper-parameter p for each layer. 

% \minjia{Why do we need to do a scaling after dropout?}
% \begin{equation}
%     R(x) = LayerNorm(M * F(x) * \frac{1}{1 - p_t} + x)
% \end{equation}

% \begin{equation}
% \begin{split}
%     X_{l+1} &= X_{l} + M_{l} \times (F(LayerNorm(X_{l}), W_{l}) + S(X_{l} + F(LayerNorm(X_{l}), W_{l}))) \times \frac{1}{1 - p_t} \\
% \end{split}
% \end{equation}

\notes{Scaling 1/1-p is similar to dropout: The outputs are scaled by a factor of $\frac{1}{1-p}$ during training. This means that during evaluation the module simply computes an identity function. In contrast, if we do not do this, then we need to scale the output during the test time by p so that a layer only contribute proportionally to the times it participates in training. More detailed discussion can be found here: \url{https://medium.com/@zhang_yang/scaling-in-neural-network-dropout-layers-with-pytorch-code-example-11436098d426}.}

\notes{
At test time all neurons are active always, which means we must scale the activations so that for each neuron: output at test time = expected output at training time. This also has the interpretation as an ensemble of all subnetworks.
Because layer drop is active only during training time but not inference time, without the scaling, the expected output would be larger during inference time because the elements are not being randomly chosen to be dropped (set to 0). But we want the expected output with and without going through the network to be the same. Therefore, during training, we compensate by making the output of the dropped layer larger by the scaling factor of 1/(1-p). A larger p means more aggressive dropout, which means the more compensation we need, i.e. the larger scaling factor 1/(1-p). Instead of making the output of the dropout layer larger during training, one could equivalently make the output of the identity function during inference smaller. However the former is easier to implement. There is a discussion on stackoverflow that provides some details. But be careful, the p in that discussion (from slides of Standford CS231n: Convolutional Neural Networks for Visual Recognition) is the ratio for keeping instead of for dropping.}

% \minjia{TODO: \sout{1. Remove my own skip-connections and extra gelu(F(x) + x) if not skipping a layer and use just the original skip connections in the BERT model. This is beneficial because they we can claim it works for all BERT-like models without huge modification or adding additional structures. 2. Try the dropout policy here.} 3. Try reduce the dropout ratio of lower-level layers (In fact, the lower layers already have a very low dropout ratio if the stack is deep, because the very first layer only has a dropout rate of 1 - 1 / L), and even the top most layer has a dropout ratio of 0.5 when we set p = 0.5.}

% \minjia{Also check Xiaodong's SAN paper. The problem he was solving was also ad-hoc. Perhaps we can stochastically drop downstream tasks as well.\minjia{Done.}}

\minjia{How does increasing network depth affect learning rate? Hank once mentioned that as the depth of the network increases, one may want to decrease the learning rate as that could mitigate the effect of large variance of gradients caused by more complex model. However, doesn't deeper model suffer from vanishing gradient issue more? To overcome vanishing gradient issue, perhaps one should consider increase the gradient?}

% To show this, we plot the FLOPS per training iteration in Fig.~\ref{fig:expected-GFLOPS}, assuming we can drop a fraction of layers at each step. Each line in the figure shows the FLOPS using different layer drop schedules. We will describe our schedule in Section~\ref{subsec:curriculum-schedule}. Regardless of which schedule to choose, the majority of FLOPS are reduced in the later steps, with the rate of reduction saturating to $\bar{\theta}$ (e.g., 0.5).

\minjia{TODO: Another option to organize the paper: Put the schedule on the right of the architecture figure. Then Remove the cosine similarity figure. Then put the similarity and vanishing gradient curve figures together. No need to put the variance curve.}

\subsection{A \Curriculum Layer Dropping Schedule}
\label{subsec:curriculum-schedule}

\minjia{TODO: more theory on curriculum learning}

% \subsubsection{Curriculum layer-drop schedule}

% \minjia{Perhaps I should think about how to combine the stochastic drop with the multi-phase idea as one instead of two separate two techniques.}`

Based on the insights from Section~\ref{subsec:training-dynamics}, and inspired by prior work on curriculum learning~\cite{curriculum-drop,curriculum-learning}
% we know that BERT training is a dynamic process and a fixed $\theta$ during training does not work well. 
\later{
using the same hyperparameters (e.g., learning rate),  which causes divergence at the beginning or at the end of training, due to the BERT training dynamics analyzed in Section~\ref{subsec:training-dynamics}. 
% In contrast, we observe that BERT pretraining is a dynamic process. 
In particular, training at the beginning is more difficult than later phases~\cite{transformer-training-tips}, due to the unbalanced gradients and unexpected large variance of gradients issue.
As a result, a warmup phase with a carefully designed learning rate schedule or a rectified Adam optimizer~\cite{radam} is often required to overcome the initial training difficulties. 
As it proceeds, the training becomes more stable and enters a cooling down phase, using a learning rate decay schedule (e.g., exponential decay).
% The warmup phase is introduced to deal with the unexpected large variance issue~\cite{radam}. 
Dropping layers at this phase is not preferred, since it would again increase the variance and hampers stability of the training. }
% \minjia{Some theoretical analysis: Why PostLN + stochastic depth causes instability (something to do with the learning rate).\minjia{We don't have theoretical explanation. Only mention empirical observation.}}
% Furthermore, we observe that lower layers of BERT model suffer more from the diminishing gradient problem. As the training proceeds, the diminishing gradients make it difficult for lower layers to learn useful representations. As a result, they share very little information with small contribution to the final goal.
% Therefore, the goal of this work is to overcome these challenges
% and to train the BERT model with as less computation as possible while maintaining the performance.
% The \gatedtransformer blocks allow training BERT with stochastic layer-wise drop reduces the expected depth of BERT training. However, using a fixed drop probability during training is a suboptimal choice.
% To tackle this issue, 
we propose a \progressiveschedule $\theta(t)$ -- a temporal schedule for the expected number of \gt blocks that are retained. 
% Intuitively, such a schedule should be high for the initial gradient updates to avoid adding additional difficulties to the warmup phase, then gradually decreasing as the training moves forward. In the late stages of training, such decrease stops.  
We limit ourselves to monotonically decreasing functions so that the likelihood of layer dropping can only increase along the temporal dimension.
% For simplicity, we use exponential decaying function $\bar{\theta}(t)$, where $t$ denotes the \emph{pacing step}. Within each step, $\bar{\theta}(t)$ remains constant for a step length $g$, which corresponds to $g$ gradient updates (e.g., g=1). 
% the  where 
% Fo which corresponds to
% % In particular, we propose 
% a temporal dependent $\bar{\theta}(t)$ parameter, where 
% t denotes the \emph{pacing step} during which $\bar{\theta}$ remains constant measured in gradient updates $t \in \{0, 1, 2, ...\}$. 
% $\bar{\theta}(t)$ models the survival probability $\bar{\theta}$ for \gt blocks at step $t$, 
We constrain ${\theta}(t)$ to be ${\theta}(t) \ge \bar{\theta}$ for any t, where $\bar{\theta}$ is a limit value, to be taken as $0.5 \le \bar{\theta} \le 0.9$ (Section~\ref{sec:eval}).
% (The training starts to diverge with $\bar{\theta} \le 0.4$ in our experiments), 
Based on this, we define the progressive schedule $\theta(t)$ as:

% \begin{defn}
% Any function $t \rightarrow \theta(t)$ such that $\theta(0)$ = 1 and $\lim_{t\rightarrow \infty}{\theta(t)} \rightarrow \bar{\theta}$ is said to be a progressive schedule with keep probability $\bar{\theta}.$
% \label{defn:schedule}
% \end{defn}

\begin{defn}
A progressive schedule is a function $t \rightarrow \theta(t)$ such that $\theta(0)$ = 1 and $\lim_{t\rightarrow \infty}{\theta(t)} \rightarrow \bar{\theta}$, where $\bar{\theta} \in (0,1]$.
\label{defn:schedule}
\end{defn}

\paragraph{Progress along the time dimension.} Starting from the initial condition $\theta(0) = 1$ where no layer drop is performed, layer drop is gradually introduced. Eventually (i.e., when $t$ is sufficiently large), $\theta(t) \rightarrow \bar{\theta}$. 
% \models the fact that the original formulation of \cite{stochastic-depth} is a special case of our schedule. 
According to Def.~\ref{defn:schedule}, in our work, we use the following schedule function:
% inspired by curriculum learning~\cite{curriculum-drop}, 
\begin{equation}
\label{eqn:curriculum-schedule}
    \bar{\theta}(t) = (1 - \bar{\theta})exp(-\gamma \cdot t) + \bar{\theta}, \gamma > 0
\end{equation}

\href{https://arxiv.org/pdf/1703.06229.pdf}{Curriculum Dropout}
% Note that \cite{curriculum-drop} is proposed for dropout at the neuron level, whereas we apply it at the \gt block level. And the original curriculum dropout is for improving model generalization and does not accelerate the training speed. 
% \minjia{Already discussed in the background.}
\later{
When t is small, the drop rate is set to zero ($\bar{\theta}(0)$ = 1) and we do not perform any \gt block drop at all. There are two reasons for this: (1) Comparing to other neural architectures, BERT pre-training is sensitive in the beginning phase, and removing the warmup stage results in serious consequences such as model divergence~\cite{transformer-training-tips,radam}.
% % due to out-of-bound variance issue and resort to a carefully designed warmup schedule to mitigate the issue~\cite{radam}.
(2) The network weights still have values which 
are close to their random and statistically independent initialization. Dropping blocks at early training steps is likely to introduce additional perturbation that makes training difficult. }

By considering Fig.~\ref{fig:expected-GFLOPS-curve-graph}, we provide intuitive and straightforward motivation for our choice. The blue curve in Fig.~\ref{fig:expected-GFLOPS-curve-graph} are polynomials of increasing degree $\delta=\{1,..,8\}$ (left to right). Despite fulfilling the initial constraint $\theta(0)=1$, they have to be manually thresholded to impose $\theta(t)\rightarrow \bar{\theta}$ when $t \rightarrow \infty$, which introduces two more parameters ($\delta$ and the threshold).
% with respect to \cite{stochastic-depth}, where the only quantity to be selected is $\bar{\theta}$. 
\later{
The yellow curve is the inverse of our proposed schedule, but it does not satisfy our initial and convergence constraints. Moreover, by evaluating the area under the curve, we can intuitively measure how much FLOPS saving it results in, which is much smaller than the proposed schedule.} 
In contrast, in our schedule, we fix $\gamma$ using the following simple heuristics $\gamma = \frac{100}{T}$, since it implies $|\theta(T) - \bar{\theta}| < 10^{-5}$ for $\theta(t) \approx \bar{\theta}$ when $t \approx T$, and it is reasonable to assume that T is at the order of magnitude of $10^4$ or $10^5$ when training Transformer networks. In other words, this means that for a big portion of the training, we are dropping ($1 - \bar{\theta}$) \gt blocks, accelerating the training efficiency. 

\minjia{Is it possible to address the warmup issue with drop schedule?}

% \minjia{Provide algorithm pseudo code}

% \subsubsection{Stochastic layer-wise drop}

\minjia{Wenhan pointed out for BERT, middle layers carry less information whereas the beginning and end layers are very important. It therefore perhaps makes sense to adjust the dropout policy to allow the end layers to participate more during the training.}

% As a result, $p$ values are set with following policy:
% We follow prior studies, where the lower the layer is, the lower we set the probability $p$. 

% \begin{itemize}
%     \item Sub-layers inside each block share the same mask, so each mask decides to drop or to keep the whole layer(including the sub-layers inside). This way we have one hyper-parameter p for each layer.
%     \item As suggested by~\cite{stochastic-depth}, the lower layers of the networks handle raw-level features and should therefore be more reliably present. 
% \end{itemize}

\paragraph{Distribution along the depth dimension.} The above \progressiveschedule assumes all gates in \gt blocks take the same $p$ value at each step t. However, as shown in Fig.~\ref{fig:similarity-analysis}, 
% higher layers are more robust to dropping than lower ones. 
% we observe that the drop of validation accuracy is more pronounced for the early \gt layers, especially in later steps (as shown in Fig.~\ref{fig:similarity-analysis}).
% setting a fixed drop rate across the entire layer hurts the convergence, presumably because dropping lower-level representations is less tolerable than dropping higher-level representations.
% As suggested by~\cite{stochastic-depth}, the 
the lower layers of the networks 
% handle raw-level features and 
should be more reliably present. Therefore, we distribute the global $\bar{\theta}$ across the entire stack so that lower layers have lower drop probability linearly scaled by their depth according to equation~\ref{eqn:linear-scale-depth}.
% with $L$ being the total number of \gt blocks. 
Furthermore, we let the sub-layers inside each block share the same schedule, so 
when $G_i$ = 1, both the inner function $f_{ATTN}$ and $f_{FFN}$ are activated, while they are skipped when $G_i$ = 0.
% ~\footnote{One possible experiment is to use a constant $p$ for all connections. However, it is possible that dropping lower-level representations is less tolerable than dropping higher-level representations}. 
Therefore, each gate has the following form during training:
\begin{equation}
\label{eqn:linear-scale-depth}
    p_l(t) = \frac{i}{L}(1 - \bar{\theta}(t))
\end{equation}

% In practice, since the amount of residual connections for transformer networks such as BERT is considerable, it is non-trivial regarding how to set the parameter $p$ for every sub-layer.
% In our design, we let the sub-layers inside each block share the same schedule, so 
% when $G_l$ = 1, both the inner function $Attn$ and $FFN$ are activated, while they are skipped when $G_l$ = 0. 

% \subsection{Interpretation}

% For a new task, we can thus simply use the off-the-shelf features of a state-of-the-art CNN pre-trained on ImageNet and train a new model on these extracted features. In practice, we either keep the pre-trained parameters fixed or tune them with a \textbf{small learning rate} in order to ensure that we do not unlearn the previously acquired knowledge. This simple approach has been shown to achieve astounding results on an array of vision tasks [13] as well as tasks that rely on visual input such as image captioning. 

% \paragraph{Implicit model ensemble}

% Some stochastic
% regularizers can make the network behave like an ensemble of networks, a pseudoensemble (Bachman et al., 2014), and can lead to marked accuracy increases. For example, the stochastic regularizer dropout creates a pseudoensemble by randomly altering some activation decisions through zero
% multiplication. 

% Averaging over many predictors leads to a reduction of the
% variance portion of the error. We present a method for evaluating the
% mean squared error of an infinite ensemble of predictors from finite (small
% size) ensemble information.

\href{https://link.springer.com/content/pdf/10.1007\%2F978-3-642-35289-8.pdf}{Large Ensemble Averaging}

\href{https://arxiv.org/pdf/1606.08415.pdf}{GAUSSIAN ERROR LINEAR UNITS (GELUS)}

\href{https://crl.ucsd.edu/~elman/Papers/elman_cognition1993.pdf}{Learning and development in neural networks : the importance of starting small}

% \subsection{Combining Stochastic and Curriculum Layer-Drop}

% \subsubsection{Expected network depth}

% In this part, we group stochastic layer-drop and Curriculum drop as a single schedule. We achieve this by setting $\theta$ to be the same in both Eqn.~\ref{eqn:linear-scale-depth} and Eqn.~\ref{eqn:curriculum-schedule}. 

Combining Eqn.~\ref{eqn:linear-scale-depth} and Eqn.~\ref{eqn:curriculum-schedule}, we have the \progressiveschedule for an \gt block below.
% and Fig.~\ref{fig:curriculum} shows the \progressiveschedule curve:
\begin{equation}
\label{eqn:combined-schedule}
    \theta_i(t) = \frac{i}{L}(1 - (1 - \bar{\theta}(t))exp(-\gamma \cdot t) - \bar{\theta}(t))
\end{equation}

% \begin{figure}[ht]
% \centering
% \includegraphics[scale=0.4]{figs/curriculum}
% \caption{\Curriculum layer-drop.}
% \label{fig:curriculum}
% \end{figure}

% Overall, we find our progressive layer drop strategy is more robust and adds improvement for the most values of $keep\_prob$.

\setlength{\intextsep}{0pt}%
\setlength{\columnsep}{5pt}%
\begin{wrapfigure}{R}{0.5\textwidth}
    \begin{minipage}{0.5\textwidth}
          \begin{algorithm}[H]
    \centering
    \vspace{0em}
    \caption{\hfill \textbf{Progressive\_\LayerDrop}}
      \begin{algorithmic}[1]
        \State \textbf{Input:} $keep\_prob$ $\bar{\theta}$
    	\label{lst:subgraph-sampling:input}
        \State InitBERT($switchable\_transformer\_block$)
        \State $\gamma\ \leftarrow \frac{100}{T}$ 
        \For {t\ $\leftarrow$ 1 to T}
            \State $\theta_t \leftarrow (1 - \bar{\theta})exp(-\gamma \cdot t) + \bar{\theta}$
            \State step $\leftarrow \frac{1 - \theta_t}{L}$
            \State $p \leftarrow 1$
            \For {l\ $\leftarrow$ 1 to L}
                % \Comment{Only showing the forward pass}
                \State action $\sim$ Bernoulli(p)
                % \State $h \leftarrow gt\_block(h)$
                \If {action == 0}
                    \State $x_{i+1} \leftarrow x_i$
                \Else
                    \State $x_{i}^{'} \leftarrow x_{i} + f_{ATTN}(f_{LN}(x_{i})) \times \frac{1}{p}$
                    \State $x_{i+1} \leftarrow x_{i}^{'} + f_{FFN}(f_{LN}(x_{i}^{'})) \times \frac{1}{p}$
                \EndIf
                \State $x_{i} \leftarrow x_{i+1}$
                \State p $\leftarrow$ p - step
            \EndFor
            \State Y $\leftarrow$ $output\_layer(x_L)$
            \State loss $\leftarrow loss\_fn(\bar{Y}, Y)$
            \State backward(loss)
        \EndFor
      \end{algorithmic}
     \label{algo:dropping-algorithm}
     \end{algorithm}
    \end{minipage}
  \end{wrapfigure}

\paragraph{Putting it together.} Note that because of the exist of the identity mapping, when an \gt block is bypassed for a specific iteration,
there is no need to perform forward-backward computation or gradient updates, and there will be updates with significantly shorter networks and more direct paths to individual layers.  
Based on this, we design a stochastic training algorithm based on \gt blocks and the \curriculum layer-drop schedule to train models like BERT faster, which we call \emph{\curriculumdrop} (Alg.~\ref{algo:dropping-algorithm}). 
\minjia{Our algorithm has no additional parameters to be tuned? No, we need to tune $\theta$.}
% As the forward-backward computation dominates the training time, dropping layers significantly speed up the training process.
The expected network depth, denoted as $\bar{L}$, becomes a random variable. Its expectation is given by: $E(\bar{L}) = \sum_{t=0}^{T}\sum_{i=1}^{L}\theta(i, t)$.
With $\bar{\theta} = 0.5$, the expected number of \gt blocks during training reduces to $E(\bar{L}) = (3L - 1)/4$ or $E(\bar{L}) \approx 3L/4$ when T is large. For the 12-layer BERT model with L=12 used in our experiments, we have $E(\bar{L}) \approx 9$. In other words, with \curriculumdrop, we train BERT with an average number of 9 layers. This significantly improves the pre-training speed of the BERT model. Following the calculations
above, approximately 25\% of FLOPS could be saved under the drop schedule with $\bar{\theta}$ = 0.5. We recover the model with full-depth blocks at fine-tuning and testing time.

\later{
% However, since during training, a sublayer $f_{ATTN}$ or $f_{FFN}$ is only active for a fraction $p_l(t)$ of all updates, and the corresponding weights
% of the next layer are calibrated for this keep probability, we re-calibrate the outputs of any given sublayer by the expected number of times it participates in training.
\minjia{This is actually problematic. In the actual code, I did not re-calibrate the output during the testing time (e.g., in validation, the stochastic flag is set to false. And if it is false, the BertEncoder will fallback to deterministic computation. Similar, in fine-tuning, although $pre\_layer\_norm$ flag is enabled. Model() in train() (line 989) does not set the stochastic depth flag to true, which means the fine-tuning is also done with stochastic depth disabled.)}

}

\section{Evaluation} 
\label{sec:eval}

% Summary of key results:
% "we are able to scale up to much larger ALBERT configurations
% that still have fewer parameters than BERT-large but achieve significantly better performance"

% "These results show that
% weight-sharing has an effect on stabilizing network parameters." 
% \minjia{We can show something similar as stochastic depth also has the effect of making training bert more stable and robust -- allows larger learning rate.}

We show that our method improves the pre-training efficiency of Transformer networks, and the trained models achieve competitive or even better performance compared to the baseline on transfer learning downstream tasks. We also show ablation studies to analyze the proposed training techniques.

% To demonstrate the effectiveness of progressive layer dropping in Transformer-based language models, we conduct experiments
% % All of our experiments are mainly with 
% on our own implementation of BERT model~\cite{bert} based on the the Huggingface[1] PyTorch implementation. 

% We evaluate on BERT~\cite{bert} to answer the following questions:
% (1) whether the proposed method can improve the BERT training efficiency at the pre-training step, and (2) whether the
% trained model can achieve similar performance compared
% to the baseline models on downstream tasks.
\later{
, and (3) whether the proposed approach is scalable as the depth of BERT increases.
}

\minjia{Train same model in less time, and train larger model with similar time.}

\minjia{Not going to the accuracy side. The goal is not to achieve the state-of-the-art accuracy results, but to explore how to make training BERT model faster.}

% Methodology
% Report the median accuracy of 5 runs for each architecture on CIFAR, reducing the impacs of random variations. 
% Report "fail" when the testing error is higher than 20\%. 
% If we find one hyperparameter is critical for training a particular type of network or setting, we conduct hyper-parameter search on X in the range of [a, b] on the training set by cross-validation. 

\minjia{Study the gradients at initialization, at output layer.}

\href{https://blog.slavv.com/37-reasons-why-your-neural-network-is-not-working-4020854bd607}{37 Reasons why your Neural Network is not working}

\minjia{Somehow it is difficult to get the same accuracy using postLN. Try preLN baseline.}

\href{https://people.eecs.berkeley.edu/~zhuohan/pdf/double-bert-icml19.pdf}{Efficient Training of BERT by Progressively Stacking}

% \minjia{This paper only evaluates on BERT-base 12-layer.}

\minjia{TODO: Calling np.random in the forward pass might be a source of slowdown. Since PyTorch might be calling the random number generator on CPU and pass the generated results to GPU}

\minjia{TODO: FP16 is easier to run into numerical instability issue. Use Fp16's validation loss. But use Fp32 for other reported results.}

\minjia{We should be able to report validation loss on seq512 when training seq128. Loading a checkpoint would definitely work (either doing it while training or after the training).}

% \subsection{Methodology}

% All of our experiments are mainly with on our own implementation of BERT model~\cite{bert} based on the the huggingface[1] PyTorch implementation. The detailed hyperparameter setting is attached in the supplementary material.  

% The model architecture (Figure 1) contains of a series of shared text encoding layers consisting of a
% lexicon encoder and a contextual transformer encoder, which are used and updated across all tasks.
% We use the huggingface[1] PyTorch implementation of BERT as a baseline model for the shared text
% encoding layers. We diverge from the MtDNN model as described by Liu et al[18] by removing
% an additional Stochastic Answering Network layer across all training tasks added in the previous
% paper, since we aim to compare the effectiveness of vanilla fine-tuning tasks as published in BERT to
% multi-task task training.

\href{https://web.stanford.edu/class/cs224n/reports/default/15734641.pdf}{Multi-Task Deep Neural Networks for Generalized Text Understanding}

\href{https://towardsdatascience.com/when-multi-task-learning-meet-with-bert-d1c49cc40a0c}{When Multi-Task Learning meet with BERT}

\href{https://github.com/OpenNMT/OpenNMT-py}{OpenNMT-py: Open-Source Neural Machine Translation}

\href{https://opennmt.net/OpenNMT-py/extended.html}{OpenNMT: Translation Example/Data}

\href{https://opennmt.net/OpenNMT-py/FAQ.html#how-do-i-use-the-transformer-model}{OpenNMT: How do I use the Transformer model?}

\paragraph{Datasets.} We follow Devlin et al.~\cite{bert} to use English Wikipedia corpus and BookCorpus for pre-training. By concatenating the two datasets, we obtain our corpus with roughly 2.8B word tokens in total, which is comparable with the data corpus used in Devlin et al.~\cite{bert}. 
\notes{Token -- Wiki: 2.04B , BC: 0.8B. }
\notes{Npy -- Wiki: 165G, BC: 58G. Val: 765M}
\notes{Text -- Wiki: 12G, BC: 4.2G. Val: 54M}
We segment
documents into sentences with 128 tokens; We normalize,
lower-case, and tokenize texts using Wordpiece tokenizer~\cite{bert}. The
final vocabulary contains 30,528 word pieces.
% Moses decoder~\cite{moses}.
% Next, we apply byte pair encoding (BPE)~\cite{bpe}. 
We split documents into one training set and one validation set (300:1). 
% The training-validation ratio for pre-training is 300:1.
% 199:1.
% 357M    /data/bert/validation_set
% 134G    /data/bert/bnorick_format/128
% 239G    /data/bert/bnorick_format/512
For fine-tuning, we use 
% We fine-tune each pre-trained model on 9 downstream tasks in
GLUE (General Language Understanding Evaluation), a collection of 9 sentence or sentence-pair natural language understanding tasks including question answering, sentiment analysis, and textual entailment. It is designed to favor sample-efficient learning and knowledge-transfer across a range of different linguistic tasks in different domains.

\paragraph{Training details.}

We use our own implementation of the BERT model~\cite{bert} based on the Huggingface[1] PyTorch implementation. All experiments are performed on 4$\times$DGX-2 boxes with 64$\times$V100 GPUs. Data parallelism is handled via PyTorch DDP (Distributed Data Parallel) library~\cite{ddp}. We recognize and eliminate additional computation overhead: we overlap data loading with computation through the asynchronous prefetching queue; we optimize the BERT output processing through sparse computation on masked tokens. Using our pre-processed data, we train a 12-layer BERT-base model from scratch as the baseline. We use a warm-up ratio of 0.02 with lr$_{max}$=1e$^{-4}$. Following \cite{bert}, we use Adam as the optimizer. We train with batch size 4K for 200K steps, which is approximately 186 epochs. The detailed parameter settings are listed in the Appendix~\ref{sec:hyperparameters}. We fine-tune GLUE tasks for 5 epochs and report the median development
set results for each task over five random initializations.

\subsection{Experimental Results}

\paragraph{Pre-training convergence comparisons.}

\begin{figure}[t]
 \centering
 \begin{minipage}[c]{0.66\textwidth}
 %\centering
  \subfloat[][\label{fig:different-lr}]{\includegraphics[scale=0.38]{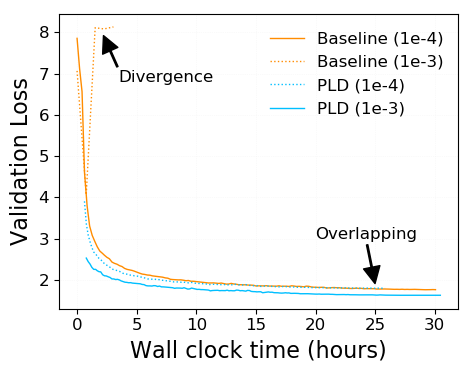}}
  \subfloat[][\label{fig:baseline-vs-pst-main}]{\includegraphics[scale=0.38]{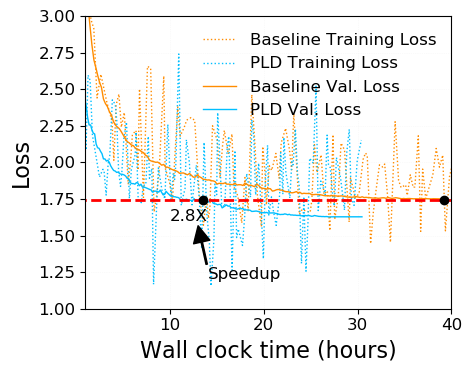}}
  \caption{The convergence curve of the baseline and our proposed method regarding the wall-clock time. }\label{fig:convergence-comparison}
 \end{minipage}%
 \hfill
 \begin{minipage}[c]{0.33\textwidth}
 %\centering
 	\newcommand{\smallcolspc}{\hspace*{0.1em}}
	\newcommand{\colspc}{\hspace*{0.28em}}
    \small
    \renewcommand\sfsmaller{}
    \centering
    \tabcolsep=0.10cm
    \captionof{table}{Training time comparison. Sample RD standards for sample reduction. SPD represents speedup.}
    \begin{tabular}{|@{\smallcolspc}c@{\smallcolspc}|@{\smallcolspc}c@{\smallcolspc}|@{\smallcolspc}c@{\smallcolspc}|@{\smallcolspc}c@{\smallcolspc}|}
\hline
 & \textbf{\begin{tabular}[c]{@{}c@{}}Training\\ Time\end{tabular}} & \textbf{\begin{tabular}[c]{@{}c@{}}Sample\\ RD\end{tabular}} & \textbf{SPD} \\ \hline
\begin{tabular}[c]{@{}c@{}}Baseline \\ ckp186\end{tabular} & 38.45h                                                           & 0                                                                   & 1                \\ \hline
    \begin{tabular}[c]{@{}c@{}}\pst \\ ckp186\end{tabular}       & 29.22h   & 0                &  1.3$\times$ \\ \hline
    \begin{tabular}[c]{@{}c@{}}\pst \\ ckp100\end{tabular}       & 15.56h   & 46\%             &  2.5$\times$ \\ \hline
    \begin{tabular}[c]{@{}c@{}}\pst \\ ckp87\end{tabular}        & 13.53h   & 53\%             &  2.8$\times$  \\ \hline
    \end{tabular}
    \label{tbl:training-time-comparison}
\end{minipage}
\end{figure}

Fig.~\ref{fig:different-lr} visualizes the convergence of validation loss regarding the computational time. We make the following observations. First, with lr$_{max}$=1e$^{-4}$, the convergence rate of our algorithm and the baseline is very close. This verifies empirically that our progressive layer dropping method does not hurt model convergence. Second, when using a larger learning rate lr$_{max}$=1e$^{-3}$, the baseline diverges. In contrast, our method shows a healthy convergence curve and is much faster. This confirms that our architectural changes stabilize training and allows BERT training with more aggressive learning rates.

% compares the validation loss of the baseline and \pst, with the same learning rates. We make the following observations. Compared with the 12-layer BERT-base baseline model (orange line), our method also reaches similar validation loss in a given amount of time. When increasing the learning rate to 1e-3, the baseline becomes numerically unstable and the training procedure diverges after consuming 8M samples, whereas our method shows a healthy convergence curve. This is because the identity mapping reordering in \gt blocks stabilizes training and allows BERT training with a larger learning rate without divergence. Fig.~\ref{fig:baseline-vs-pst-main} shows both the training curve and the validation curve of the baseline and the model trained with \pst. \pst has a much faster convergence speed than the baseline, and to reach the same validation loss (e.g., 1.77), \pst is 2.77 times faster. 

\paragraph{Speedup.}

Fig.~\ref{fig:baseline-vs-pst-main} shows both the training curve (dotted) and the validation curve (solid) of the baseline and \pst with a zoomed-in view. The baseline curve becomes almost flat at epoch 186, getting a validation loss of 1.75. In contrast, \pst reaches the same validation loss at epoch 87, with 53\% fewer training samples. Furthermore, \pst achieves a 24\% time reduction when training the same number of samples. This is because our approach trains the model with a smaller number of expected depth for the same number of steps. It is slightly lower than the 25\% GFLOPS reduction in the analysis because the output layer still takes a small amount of computation even after optimizations. The combination of these two factors, yields 2.8$\times$ speedup in end-to-end wall-clock training time over the baseline, as shown in Table~\ref{tbl:training-time-comparison}.

\minjia{Not going to the accuracy side. The goal is not to achieve the state-of-the-art accuracy results, but to explore how to make training BERT model faster.}

\paragraph{Downstream task accuracy.}

Despite improved training speed, one may still wonder whether such a method is as effective as the baseline model on downstream tasks. 
% We compare the performance of the model trained with \pst and the baseline on GLUE tasks. 
Table~\ref{tbl:downstream-accuracy} shows our results on the GLUE dataset compared to the baseline. Our baseline is comparable with the original BERT-Base (on the test set), and our \pst method achieves a higher GLUE score than our baseline (83.2 vs. 82.1) when fine-tuning the checkpoint (186). We also dump model checkpoints from different epochs during pre-training and fine-tune these models.
The checkpoint 87 corresponds to the validation loss at 1.75 achieved by \pst. The GLUE score is slightly worse than the baseline (81.6 vs. 82.1). However, by fine-tuning at checkpoint 100, \pst achieves a higher score than the baseline (82.3 vs. 82.1) at checkpoint 186. In terms of the pre-training wall clock time, \pst requires 15.56h vs. the baseline with 39.15h to get similar accuracy on downstream tasks, which corresponds to a 2.5$\times$ speedup.
% We also compare BERT-base models with state-of-the-art pre-BERT models on the GLUE lead board: OpenAI GPT-2~\cite{gpt-2} and ELMo~\cite{ELMo}. 

% algorithm is faster and achieves competitive performance comparing to the baseline models on various tasks.
\minjia{TODO: Add footnotes that GLUE leader board is no longer useful.}

\begin{table}[!ht]
	\newcommand{\smallcolspc}{\hspace*{0.06em}}
	\newcommand{\colspc}{\hspace*{0.28em}}
    \small
    \renewcommand\sfsmaller{}
    \centering
\tabcolsep=0.10cm
    \caption{The results on the GLUE benchmark. The number below each task denotes the number of training examples.
    The metrics for these tasks can be found in the GLUE paper~\cite{glue}. We compute the geometric mean of the metrics as the GLUE score.}
\begin{tabular}{|@{\smallcolspc}l@{\smallcolspc}|@{\smallcolspc}c@{\smallcolspc}|@{\smallcolspc}c@{\smallcolspc}|@{\smallcolspc}c@{\smallcolspc}|@{\smallcolspc}c@{\smallcolspc}|@{\smallcolspc}c@{\smallcolspc}|@{\smallcolspc}c@{\smallcolspc}|@{\smallcolspc}c@{\smallcolspc}|@{\smallcolspc}c@{\smallcolspc}|l|}
\hline
\multirow{2}{*}{Model}      & \begin{tabular}[c]{@{}l@{}}RTE\\   (Acc.)\end{tabular} & \begin{tabular}[c]{@{}l@{}}MRPC\\   (F1/Acc.)\end{tabular} & \begin{tabular}[c]{@{}l@{}}STS-B\\   (PCC/SCC)\end{tabular} & \begin{tabular}[c]{@{}l@{}}CoLA \\   (MCC)\end{tabular} & \begin{tabular}[c]{@{}l@{}}SST-2\\   (Acc.)\end{tabular} & \begin{tabular}[c]{@{}l@{}}QNLI\\   (Acc.)\end{tabular} & \begin{tabular}[c]{@{}l@{}}QQP\\   (F1/Acc.)\end{tabular} & \begin{tabular}[c]{@{}l@{}}MNLI-mm\\-/m   (Acc.)\end{tabular}  & \multicolumn{1}{c|}{\multirow{2}{*}{GLUE}} \\ \cline{2-9} 
                                        & 2.5K                                                   & 3.7K                                                       & 5.7K                                                        & 8.5K                                                    & 67K                                                      & 108K                                                    & 368K                                                      & 393K   & \multicolumn{1}{c|}{}                      \\ \hline
% ELMo-BiLSTM-Attn                        & 58.9       & 84.4/78.0      & 74.2/72.3       & 33.6       & 90.4         & 74.5        & 63.1/84.3     & 74.1/74.5        & 70.0                                \\ \hline
% OpenAI GPT                              & \textbf{69.1}       & 87.7/83.7      & 85.3/84.8       & 47.2       & 93.1         & 80.7        & 70.1/88.1     & 80.7/80.6        & 78.2                                \\ \hline
BERT$_{base}$ (original)                    & 66.4       & 88.9/84.8      & 87.1/89.2       & 52.1       & \textbf{93.5}         & \textbf{90.5}        & 71.2/89.2     & \textbf{84.6}/83.4        & 80.7                                \\ \hline
% BERT$_{base}$  (baseline)                    & 67.8       & 88.0/\textbf{86.0}      & 89.5/\textbf{89.2}       & 52.5       & 91.2         & 87.1        & 89.0/90.6     & 82.5/83.4        & 82.1                                \\ \hline
% BERT$_{base}$  (\pst) & \textbf{69}         & 87.2/85.1      & \textbf{89.6}/89.1       & \textbf{59.4}       & 91.8         & 88          & \textbf{89.4}/\textbf{90.9}     & 83.1/\textbf{83.5}        & \textbf{83.2}                                \\ \hline
% \end{tabular}
% BERT-base   (original)            & 66.4                  & \textbf{88.9/84.8}        & 87.1/89.2                  & 52.1          & \textbf{93.5}           & \textbf{90.5}          & 71.2/89.2                & \textbf{84.6/83.4}          & 80.7          \\ \hline
BERT$_{base}$ (Baseline, ckp186)            & 67.8                  & 88.0/86.0                 & 89.5/\textbf{89.2}                  & 52.5          & 91.2                    & 87.1                   & 89.0/90.6                & 82.5/83.4                   & 82.1          \\ \hline
BERT$_{base}$ (\pst, ckp87)           & 66                    & 88.2/85.6                 & 88.9/88.4                  & 54.5 & 91                      & 86.3         & 87.4/89.1       & 81.6/82.4                   & 81.6          \\ \hline
BERT$_{base}$ (\pst, ckp100) & 68.2                  & 88.2/85.8                 & 89.3/88.9                  & 56.1          & 91.5                    & 86.9                   & 87.7/89.3                & 82.4/82.6                   & 82.3          \\ \hline
BERT$_{base}$ (\pst, ckp186)         & \textbf{69}           & \textbf{88.9/86.5}        & \textbf{89.6}/89.1         & \textbf{59.4} & 91.8                    & 88                     & \textbf{89.4/90.9}       & 83.1/\textbf{83.5}                   & \textbf{83.2} \\ \hline
\end{tabular}
\minjia{It is weird to report just Squad. Also, GLUE release test. So no longer need to report test results. Dev results are sufficient. Need to update the other model's results. May not need ELMo and OpenAI.}
% \minjia{For STS-B, we report Pearson and Spearman correlation.}
\label{tbl:downstream-accuracy}
\end{table}

\begin{figure}[ht!]
\centering
\small
\subfloat[MNLI-m]{\includegraphics[scale=0.33, keepaspectratio=true]{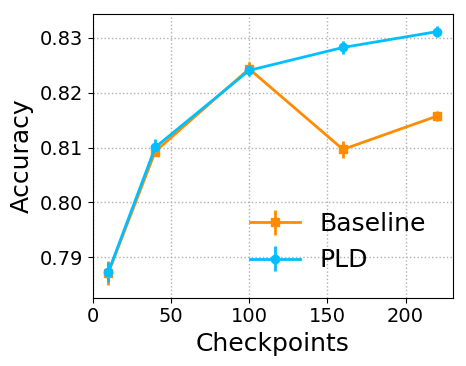}\label{fig:MNLI-m-comparison}}
\subfloat[MNLI-mm]{\includegraphics[scale=0.33, keepaspectratio=true]{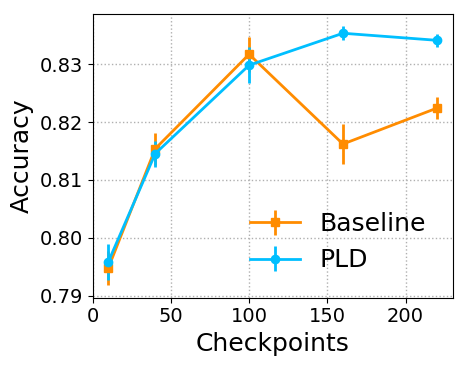}\label{fig:MNLI-mm-comparison}}
\subfloat[QNLI]{\includegraphics[scale=0.33, keepaspectratio=true]{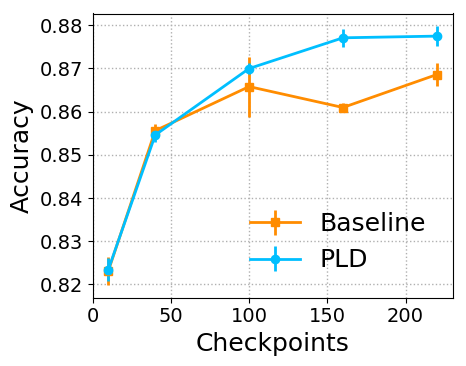}\label{fig:QNLI-comparison}} \\
\subfloat[QQP]{\includegraphics[scale=0.33, keepaspectratio=true]{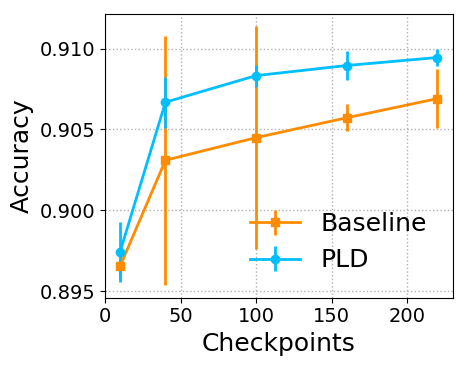}\label{fig:QQP-comparison}} 
\subfloat[RTE]{\includegraphics[scale=0.33, keepaspectratio=true]{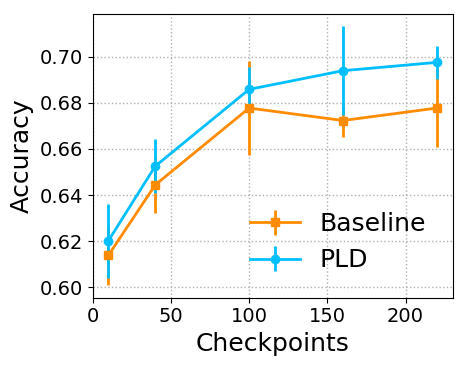}\label{fig:RTE-comparison}}
\subfloat[SST-2]{\includegraphics[scale=0.33, keepaspectratio=true]{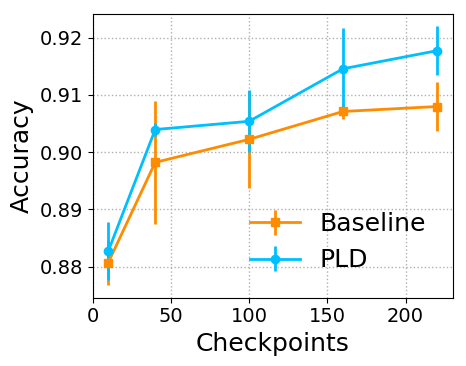}\label{fig:SST-2-comparison}} \\
\subfloat[WNLI]{\includegraphics[scale=0.33, keepaspectratio=true]{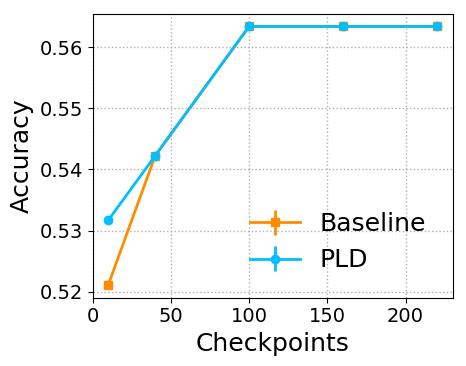}\label{fig:WNLI-comparison}} 
\subfloat[CoLA]{\includegraphics[scale=0.33, keepaspectratio=true]{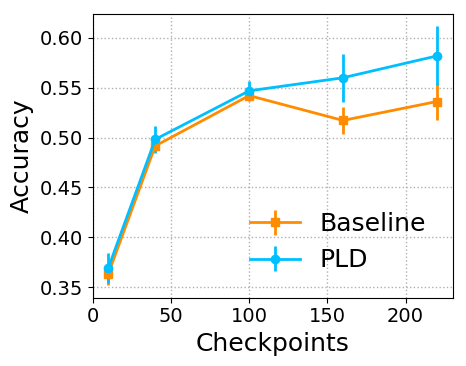}\label{fig:CoLA-comparison}}
\subfloat[MRPC (acc.)]{\includegraphics[scale=0.33, keepaspectratio=true]{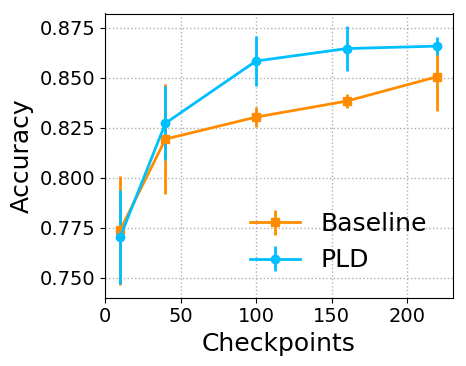}\label{fig:MRPC-acc-comparison}} \\
\subfloat[MRPC (F1.)]{\includegraphics[scale=0.33, keepaspectratio=true]{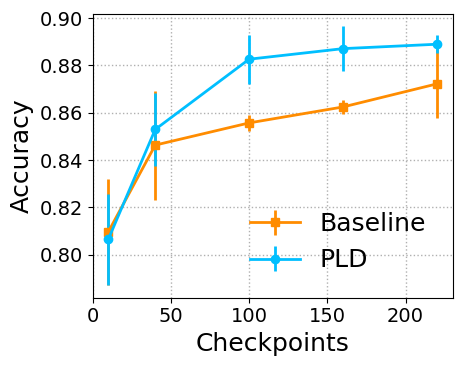}\label{fig:MRPC-f1-comparison}}
\subfloat[SST-B (PCC)]{\includegraphics[scale=0.33, keepaspectratio=true]{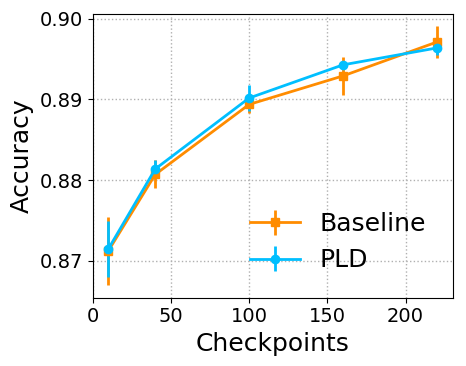}\label{fig:SST-B-pearson-comparison}}
\subfloat[SST-B (SCC)]{\includegraphics[scale=0.33, keepaspectratio=true]{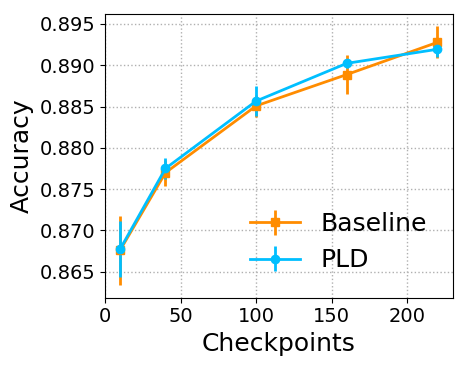}\label{fig:SST-B-spearman-comparison}}
\caption{The fine-tuning results at different checkpoints.}
\label{fig:fine-tune-comparison}
\end{figure}

Fig.~\ref{fig:fine-tune-comparison} illustrates the fine-tuning results between the baseline and \pst on GLUE tasks over different checkpoints
% We dump models from different checkpoints during pre-training and fine-tune them on downstream tasks. 
Overall, we observe that \pst not only trains BERT faster in pre-training but also preserves the performance on downstream tasks. In each figure, we observe that both curves have a similar shape at the beginning because no layer drop is added. For later checkpoints, \pst smoothly adds layer drop. Interestingly, we note that the baseline model has fluctuations in testing accuracy. In contrast, the downstream task accuracy from \pst is consistently increasing as the
number of training epochs increases. This indicates that \pst takes a more robust optimization path toward the optimum. We also observe that our model achieves higher performance on MNLI, QNLI, QQP, RTE, SST-2, and CoLA on later checkpoints, indicating that the model trained with our approach also generalizes better on downstream tasks than our baseline does.  

% In each figure, both curves have a similar shape at the beginning, because no layer drop is added in the beginning. For later checkpoints, \pst smoothly adds layer drop. Although \pst takes shorter time to train, it keeps competitive performance comparing to the baseline on the GLUE tasks. This fact confirms the validity of \pst. 

% Furthermore, the default BERT model has fluctuations in testing accuracy, whereas the downstream task accuracy from \pst is consistently increasing as the
% number of training epochs increases. This indicates that \pst takes a more robust path toward the optimum and has better
% generalization performance along the optimization path.
From a knowledge transferability perspective, the goal of training a language model is to learn a good representation of natural language that ideally ignores the data-dependent noise and generalizes well to downstream tasks. However, training a model with a constant depth is at least somewhat noisy and can bias the model to prefer certain representations, whereas \pst enables more sub-network configurations to be created during training Transformer networks.
Each of the L \gt blocks is either active or inactive, resulting in $2^L$ possible network combinations. By selecting a different submodular in each mini-batch, \pst encourages the submodular to produce good results independently. This allows the unsupervised pre-training model to obtain a more general representation by averaging the noise patterns, which helps the model to better generalize to new tasks. 
On the other hand, during inference, the full network is presented, causing the effect of ensembling different sub-networks.

% On the other hand, 
% during inference the full network is presented, causing the effect of ensembling different sub-networks.
% keeps competitive or sometimes higher performance comparing to the baseline on the GLUE tasks.

\minjia{Original explanation in "ResNet as ensemble" uses "short path" and "long path" to explain why ResNet achieves good results. May need to understand what path refers to.}

\href{https://arxiv.org/pdf/1712.03556.pdf}{Stochastic Answer Networks for Machine Reading Comprehension}

\subsection{Ablation Studies}
\label{subsec:ablation}

\paragraph{Downstream task fine-tuning sensitivity.}

\begin{figure}[ht!]
\centering
\small
\subfloat[MNLI-m]{\includegraphics[scale=0.33, keepaspectratio=true]{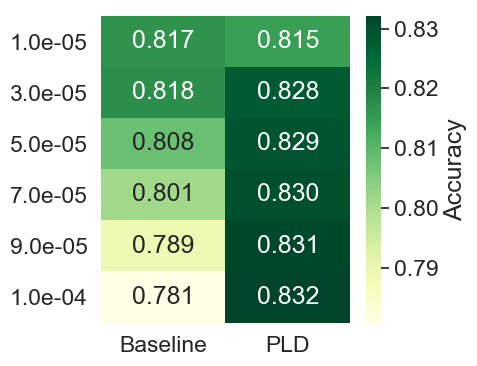}\label{fig:heatmap-MNLI-m}}
\subfloat[MNLI-mm]{\includegraphics[scale=0.33, keepaspectratio=true]{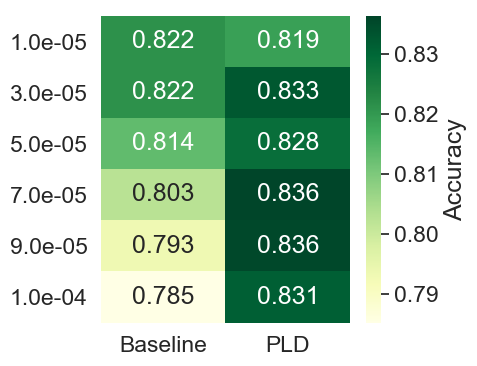}\label{fig:heatmap/heatmap-MNLI-mm}}
\subfloat[QNLI]{\includegraphics[scale=0.33, keepaspectratio=true]{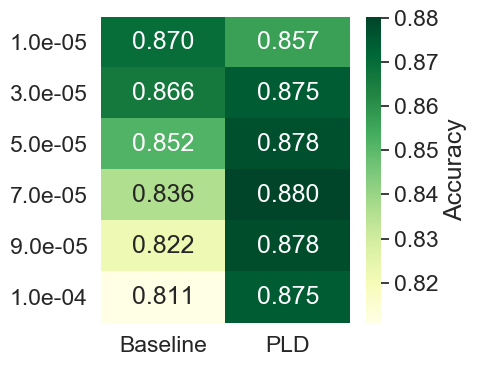}\label{fig:heatmap/heatmap-QNLI}} \\
\subfloat[QQP]{\includegraphics[scale=0.33, keepaspectratio=true]{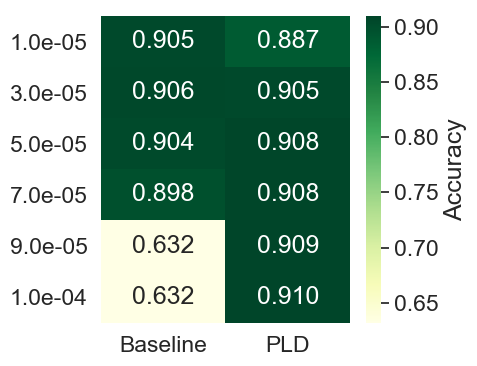}\label{fig:heatmap/heatmap-QQP}} 
\subfloat[RTE]{\includegraphics[scale=0.33, keepaspectratio=true]{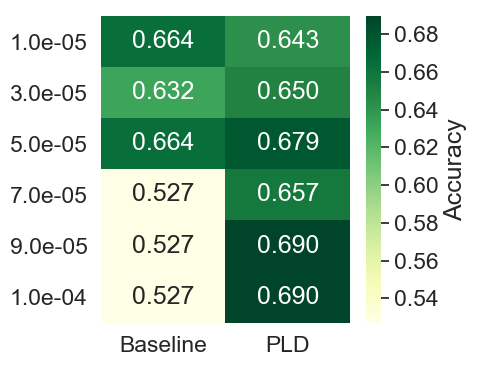}\label{fig:heatmap/heatmap-RTE}}
\subfloat[SST-2]{\includegraphics[scale=0.33, keepaspectratio=true]{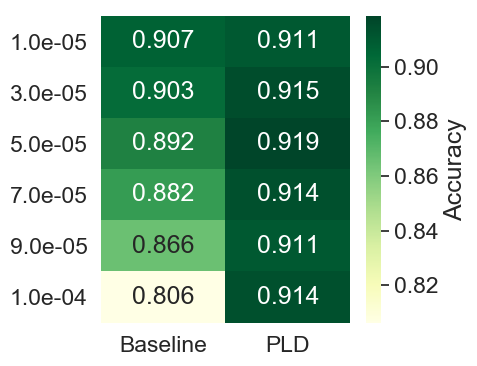}\label{fig:heatmap/heatmap-SST-2}} 
\caption{The fine-tuning results at different checkpoints.}
\label{fig:fine-tune-heatmap}
\end{figure}

To further verify that our approach not only stabilizes training but also improves downstream tasks, we show a grid search on learning rates 
\{1e-5, 3e-5, 5e-5, 7e-5, 9e-5, 1e-4\}. As illustrated in Fig.~\ref{fig:fine-tune-heatmap}, the baseline is vulnerable to the choice of learning rates. Specifically, the fine-tuning results are often much worse with a large learning rate, while \pst is more robust and often achieves better results with large learning rates. 

\paragraph{The Effect of $\bar{\theta}$.} We test different values of the keep ratio $\bar{\theta}$ and identify $0.5 \le \bar{\theta} \le 0.9$ as a good range, as shown in Fig.~\ref{fig:varying-keep-ratio} in the Appendix. We observe that the algorithm may diverge if $\bar{\theta}$ is too small (e.g., 0.3).   

\begin{figure}
    \centering
    \includegraphics[scale=0.5]{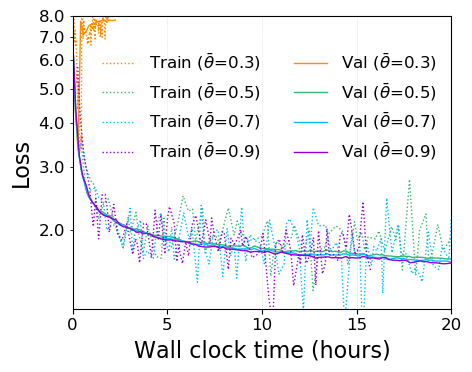}
    \caption{Convergence curves varying the keep ratio $\bar{\theta}$.}
    \label{fig:varying-keep-ratio}
\end{figure}

\paragraph{\pst vs. PreLN.} To investigate the question on how \pst compares with PreLN, we run both PreLN with the hyperparameters used for training PostLN (lr=1e-4) and the hyperparameters used for \pst (lr=1e-3) to address the effect from the choice of hyperparameters. We train all configurations for the same number of epochs and fine-tune following the standard procedure. In both cases, PreLN is 24\% slower than \pst, because PreLN still needs to perform the full forward and backward propagation in each iteration.

Table~\ref{tbl:downstream-accuracy-ablation} shows the fine-tuning results on GLUE tasks. When trained with the same hyperparameters as PostLN, PreLN appears to have a much worse GLUE score (80.2) compared with PostLN (82.1) on downstream tasks. This is because PreLN restricts layer outputs from depending too much on their own residual branches and inhibits the network from reaching its full potential, as recently studied in \cite{understanding-transformer-difficulty}. When trained with the large learning rate as \pst, PreLN's result have improved to 82.6 but is 0.6 points worse than \pst (83.2), despite using 24\% more compute resource. \pst achieves better accuracy than PreLN because it encourages each residual branch to produce good results independently. 

\begin{table}[!ht]
	\newcommand{\smallcolspc}{\hspace*{0.05em}}
	\newcommand{\colspc}{\hspace*{0.28em}}
    \small
    \centering
\tabcolsep=0.10cm
\caption{Ablation studies of the fine-tuning results on the GLUE benchmark.}
\begin{tabular}{|@{\smallcolspc}l@{\smallcolspc}|@{\smallcolspc}l@{\smallcolspc}|@{\smallcolspc}l@{\smallcolspc}|@{\smallcolspc}l@{\smallcolspc}|@{\smallcolspc}l@{\smallcolspc}|@{\smallcolspc}l@{\smallcolspc}|@{\smallcolspc}l@{\smallcolspc}|@{\smallcolspc}l@{\smallcolspc}|@{\smallcolspc}l@{\smallcolspc}|@{\smallcolspc}l@{\smallcolspc}|}
\hline
Model    & \begin{tabular}[c]{@{}l@{}}RTE\\   (Acc.)\end{tabular} & \begin{tabular}[c]{@{}l@{}}MRPC\\   (F1/Acc.)\end{tabular} & \begin{tabular}[c]{@{}l@{}}STS-B\\   (PCC/SCC)\end{tabular} & \begin{tabular}[c]{@{}l@{}}CoLA \\   (MCC)\end{tabular} & \begin{tabular}[c]{@{}l@{}}SST-2\\   (Acc.)\end{tabular} & \begin{tabular}[c]{@{}l@{}}QNLI\\   (Acc.)\end{tabular} & \begin{tabular}[c]{@{}l@{}}QQP\\   (F1/Acc.)\end{tabular} & \begin{tabular}[c]{@{}l@{}}MNLI-m/mm\\   (Acc.)\end{tabular}  & GLUE \\ \hline
BERT   (Original)            & 66.4                  & \textbf{88.9}/84.8        & 87.1/89.2                  & 52.1          & \textbf{93.5}           & \textbf{90.5}          & 71.2/89.2                & \textbf{84.6/83.4}          & 80.7          \\ \hline
BERT + PostLN            & 67.8                  & 88.0/86.0                 & 89.5/89.2                  & 52.5          & 91.2                    & 87.1                   & 89.0/90.6                & 82.5/83.4                   & 82.1          \\ \hline
BERT + PreLN + Same lr    &66.0	    & 85.9/83.3	& 88.2/87.9	& 46.4	& 90.5	& 85.5    &	89.0/90.6	& 81.6/81.6
                  & 80.2          \\ \hline
BERT + PreLN + lr$\uparrow$    &67.8	    & 86.7/84.5	& \textbf{89.6/89.1}	& 54.6	& 91.9	& 88.1    &	89.3/\textbf{90.9}	& \textbf{83.6/83.7}
                  & 82.6          \\ \hline
Shallow BERT + PreLN + lr$\uparrow$    &66.0	    & 85.9/83.5	& 89.5/88.9	& 54.7	& 91.8	& 86.1    &	89.0/90.6	& 82.7/82.9
                  & 81.8          \\ \hline
BERT + PreLN + lr$\uparrow$ + Rand.   
                   &68.2	    & 88.2/86.2	& 89.3/88.8	& 56.8	& 91.5	& 87.2    &	88.6/90.3	& 82.9/83.3
                  & 82.7          \\ \hline
BERT + PreLN + lr$\uparrow$ + TD   
                  &68.2	    & 88.6/\textbf{86.7}	& 89.4/88.9	& 55.9	& 91.3	& 86.8    &	89.1/90.7	& 82.7/83.1
                  & 82.7          \\ \hline
% BERT + PreLN + Large lr + DD only     &66	    & 85.9/83.5	& 89.5/88.9	& 54.7	& 91.8	& 86.1    &	89.0/90.6	& 82.7/82.9
%                   & 81.8          \\ \hline
BERT + PreLN + lr$\uparrow$ + PLD         & \textbf{69.0}           & \textbf{88.9}/86.5        & \textbf{89.6/89.1}         & \textbf{59.4} & 91.8                    & 88.0                     & \textbf{89.4/90.9}       & 83.1/83.5                   & \textbf{83.2} \\
\hline
\end{tabular}
\label{tbl:downstream-accuracy-ablation}
\end{table}

\paragraph{\pst vs. Shallow network.} 
% We find that directly training 9-layer BERT (BERT-L9), which has the same number of computational GFLOPS as ours, leads to suboptimal performance on the downstream task accuracy, shown as \emph{Shallow BERT + PreLN + Large lr} in Table~\ref{tbl:downstream-accuracy-ablation}.
% the validation set, as shown in Fig.~\ref{fig:comparison-to-reduced-depth}. 
\emph{Shallow BERT + PreLN + Large lr} in Table~\ref{tbl:downstream-accuracy-ablation} shows the downstream task accuracy of the 9-layer BERT. Although having the same same number of training computational GFLOPS as ours, the shallow BERT
% directly trains a 9-layer BERT without sampling. This configuration 
underperforms PreLN by 0.8 points and is 1.4 points worse than \pst likely because the model capacity has been reduced by the loss of parameters. 
% presumably because of insufficient model complexity for capturing textual variations and distributional similarity. 

\paragraph{\pst vs. Random drop.} 
% Similar to Stochastic Depth~\cite{stochastic-depth}, if we set a fixed layer drop rate (0.75) during the entire training, the generalization gap is large, as shown in Fig.~\ref{fig:comparison-to-stochastic-depth}. 
\emph{BERT + PreLN + Large lr + Random} drops layers randomly with a fixed ratio (i.e., it has the same compute cost but without any schedule), similar to Stochastic Depth~\cite{stochastic-depth}. The GLUE score is 0.9 points better than shallow BERT under the same compute cost and 0.1 point better than PreLN while being 24\% faster, indicating the strong regularization effect from stochastic depth. It is 0.5 points worse than \pst, presumably because a fixed ratio does not take into account the training dynamics of Transformer networks. 
% which demonstrates the benefit of having schedules enabled. 
% which indicates that stochastic depth is not directly applicable to language models.

\paragraph{Schedule impact analysis.} \emph{BERT + PreLN + Large lr + TD only (32-bit*)} disables the schedule along the depth dimension (DD) and enables only the schedule along the temporal dimension (TD) in training. Its GLUE score matches "Random", suggesting that the temporal schedule has similar performance as the fixed constant schedule along the time dimension and accuracy gains of \pst is mostly from the depth dimension. However, without the temporal schedule enabled, the model diverges with \emph{NaN} in the middle of half-precision (16-bit) training and has to switch to full-precision (32-bit) training, slowing down training speed. Furthermore, this concept of starting-easy and gradually increasing the difficulty of the learning problem has its roots in curriculum learning and often makes optimization easier. We adopt the temporal schedule since it is robust and helpful for training stability, retaining similar accuracy while reducing training cost considerably.

\notes{
\begin{itemize}
    \item Compared with the 12-layer BERT-base baseline model (orange line), our method also reaches similar pre-training training loss and validation loss at the end of training. However, the training time of our proposed method is about 15\% shorter (XXX hours vs. XXX hours). This is mainly because for the same number of steps, our approach trains the model with a smaller number of expected depth. It is lower than the 25\% speedup because there are additional layers (output layers) that take a large amount of time. In the scalability section, we show that as the depth of the BERT model increases, the speedup becomes more.

    \item The switch paradigm induces a discontinuity in the objective value which can damage the performance with respect to the smooth transition performed by our \progressiveschedule.
    
    % \item  Anti-curriculum performs slightly better or worse than the no-curriculum strategy and always worse than any curriculum implementation. However, it does not provide much training speedup because XXX
    % \minjia{This seems to indicate that anti-curriculum learning always make the results worse}
    \minjia{Anti-curriculum is one that is difficult to argue. Because it should provide some speedup. Given that the exact speedup is difficult to quantify without giving a schedule. It might be better not to mention.}

    \item \Curriculum layer-drop smoothly increases the drop rate as training evolves, thereby improving the generalization of the model.
    
    \item Compared with other layer drop training algorithms (regular and switch), our method can achieve either lower loss or faster speed. The other algorithms do not take training difficulties at different phases and depths into consideration, so they converge slower.
\end{itemize}
}

% \subsection{Comparison to directly train a reduced layer BERT.}
% \subsection{Comparison to Other Approaches}

% \subsection{How does the training time compare?}

% \input{related}

\section{Conclusion}
\label{sec:conclusion}

Unsupervised language model pre-training is a crucial step for getting state-of-the-art performance on NLP tasks. The current time for training such a model is excruciatingly long, and it is very much desirable to reduce the turnaround time for training such models. In this paper, we study the efficient training algorithms for pre-training BERT model for NLP tasks. We have conducted extensive analysis and found that model architecture is important when training Transformer-based models with stochastic depth. Using this insight, we propose the \gatedtransformer block and a progressive layer-wise drop schedule. Our experiment results show that our training strategy achieves competitive performance to training a deep model from scratch at a faster rate.

% \paragraph{TODO:}

\later{
\begin{itemize}
    \item Test out GPT-2. Then this is no longer an ad hoc one.
    \item Can I prove "This verifies our theoretical analysis that the proposed algorithm converges to critical points with a rate of O(1/T). -- ouroboros"
    \item how to prove mathematically our approach (ensemble/stochastic depth) is more beneficial to downstream tasks. We need some theoretical analysis and explanation. 
    \item The story should go this way: Accelerating BERT is crucial. The BERT computation is linear wrt to the Transformer blocks. Straightforward idea is to reduce the depth. However, reduced depth causes serious issues such as low accuracy. On the other hand, stochastic depth is difficult to train, because XXX and it does not give on-par accuracy (show training curve). Given that the depth of the model is reduced, if we want to make the training converge faster, increase the learning rate, the training simply diverges (given analytical results and experiment results). Then we give our solution: (1) switchable Transformer blocks make training stable and converge faster. (2) Progressive layer drop schedule to overcome warmup. Need some deeper understanding of stochastic depth.
    \minjia{Also add that stochastic depth + postLN leads to suboptimal results, because we cannot use a large learning rate.}
    \item We need theoretical analysis (find some followup of stochastic depth and curriculum learning -- add theorem and proofs -- prove that stochastic depth works, and prove that curriculum/progressive schedule is correct (converge) -- essentially deeper explanation).
    \item {TODO: Add another subsection in the analysis part on how removing layers would have an impact to model training dynamics. Similar as Section 4.1 in https://arxiv.org/pdf/1605.06431.pdf.}
    \item Show out-of-range variance figure (in motivation). This can be done by setting warmup ratio as 0 so training directly start with high learning rate.
    \item Add results on regular drop rate: 0.75 is fine, but how about 0.5? This is helpful to show a constant drop rate is suboptimal (we need this claim).
    \item Try different keep ratio.
    \item Fine-tune 9-layer BERT -- change the finetune script to get 9 layer.
    \item Try switch-curriculum.
    \item Show variance analysis for preLN and postLN, similar to Xiaodong's work.
    \item Adjust learning rate dynamically based on the stochastic depth?
    \item Adjust curriculum based on the learning rate
        \minjia{The issue is that it then becomes coupled with the learning rate schedule, We need something more general than the curriculum curve we provided.}
    \item \sout{May need to redraw training curve and validation curve based on the updated model performance results.}
   \item  \sout{If we are going to show our approach with two curves with different lr, we probably should do the same for the baseline.}
    \item \sout{Try again on seq512 using a larger lr to get the validation curve} 
    \item \sout{May need to switch the data preprocessing pipeline.}
    \item \sout{Investigate whether seq128 can show validation curve.}
    \item \sout{Train with a 9-layer BERT directly.}
    \item \sout{May need to report fine-tune results.}
    \item \sout{Investigate why actual training takes longer than expected. Using instrumentation.}
    \item \sout{Change data loader worker count to 1 and measure all drop performance again.}
    \item \sout{Report variance of fine-tuning results (right methodology). }
    \item \sout{Vanishing gradients for 24-layer, 48-layer BERT. There is a caveat, although postLN initially has vanishing gradient problem, it later disappears after a few hundred steps. In the "On the Pre-Layer Norm" paper, they mentioned "the gradients are well behaved without any exploding or vanishing \textbf{at initialization} for the Pre-LN Transformer both theoretically and empirically". So this is an effect happens only at the initial stage. If it happens only at the initial stage, then it is possible to address it through good initialization. }
    \item \sout{Plot more than one step of gradient information (only if it helps)}
    \item \sout{Plot different learning rate / differences of fine-tune results}
    \item \sout{Fine-tune baseline and report variance.}
    \item \sout{Collect full fine-tuning results. May need to be based on the one with the larger learning rate (only our results need to be updated. The baseline one can still use their best results).}
\end{itemize}
}
% \begin{itemize}
%     \item Adjust learning rate dynamically based on the stochastic depth?
% \end{itemize}

\minjia{Niranjan recently demonstrated that it was possible to train Adam with 64K batch size and PreLN. This confirms the understanding here. By incorporating this finding, we can come up with a paper that talks about large batch training + Adam + Stochastic depth + ZeRO.  I think some learnings are (1) PreLn enables Adam to scale to large batch (PostLn Adam hits 4k but not beyond) (2) If Lamb can be avoided, Adam instead has lesser hyperparameters to tune (no min and max) and also Adam is less expensive than Lamb due to to no normalization; these 2 learnings can then be used for large model (eg, Turing NLG, etc); perhaps there are more insight as well?}

% \subsection{Implementation}

% Generating stochastic numbers in the forward pass may cause two sources of overhead:

% (1) Allocating new tensors in the forward pass may incur additional overhead if we need to pass the generated number from CPU to GPU.

% (2) We include additional control flow that may hurt the performance of PyTorch (not a common case).
% To verify this one, I should first teset it on CPU.

% According to our theoretical analysis, when putting the
% layer normalization between the residual blocks, the \textbf{expected gradients of the parameters near the
% output layer are large}. I observe this effect as well, but how to prove that from theoretical perspective?

% Given the \textbf{gradients are well-behaved} in the PreLN Transformer, it is natural to consider removing the learning rate warm-up stage during
% training.

% By studying the \textbf{gradients at initialization}, we
% show why the learning rate warm-up stage is essential in training the Post-LN Transformer. 

\medskip

% \clearpage
% \setcounter{page}{1}

% \input{broader-impact}

% \input{ack}

{\small \bibliographystyle{unsrt}
\bibliography{reference}} 

\newpage

\appendix

\section{Pre-training Hyperparameters}
\label{sec:hyperparameters}

Table~\ref{tbl:pretraining-hyperparameters} describes the hyperparameters for pre-training the baseline and \pst. 

\begin{table}[!ht]
    \centering
    \caption{Hyperparameters for pre-training the baseline and \pst.}
\begin{tabular}{|l|l|l|}
\hline
\textbf{Hyperparameter}              & \textbf{Baseline}          & \textbf{\pst}             \\ \hline
Number of   Layers               & 12                         & 12                         \\ \hline
Hidden   zies                    & 768                        & 768                        \\ \hline
Attention   heads                & 12                         & 12                         \\ \hline
Dropout                          & 0.1                        & 0.1                        \\ \hline
Attention   dropout              & 0.1                        & 0.1                        \\ \hline
Total batch size                 & 4K                         & 4K                         \\ \hline
Train micro batch   size per gpu & 16                         & 16                         \\ \hline
Optimizer                        & Adam                       & Adam                       \\ \hline
Peak learning rate               & 1e-04                   & 1e-03                   \\ \hline
Learning rate scheduler          & warmup\_linear\_decay\_exp & warmup\_linear\_decay\_exp \\ \hline
Warmup ratio                     & 0.02                       & 0.02                       \\ \hline
Decay rate                       & 0.99                       & 0.99                       \\ \hline
Decay step                       & 1000                       & 1000                       \\ \hline
Max Training steps               & 200000                     & 200000                     \\ \hline
Weight   decay                   & 0.01                       & 0.01                       \\ \hline
Gradient   clipping              & 1                          & 1                          \\ \hline
\end{tabular}
\label{tbl:pretraining-hyperparameters}
\end{table}

\section{Establishing Identity Mapping with PreLN}
\label{sec:preln-analysis}

Prior studies~\cite{resnet,identity-mapping} suggest that establishing \emph{identity mapping} to keep a \emph{clean} information path (no operations except addition) is helpful for easing optimization of networks with residual connections. With the change of PreLN, we can express the output of the i-th Transformer layer as the input $x_i$ of that layer plus a residual transformation function $f_{RT} = f_{S-ATTN}(f_{LN}(x_i)) + f_{FFN}(f_{LN}(x_i^{'}))$, and the output layer $x_L =  x_l + \sum_{i=l}^{L-1} f_{RT}(x_i)$ as the recursive summation of preceding $f_{RT}$ functions in shallower layers (plus $x_l$). If we denote the loss function as $\mathcal{E}$, from the chain rule of backpropagation~\cite{auto-diff-survey} we have:

\begin{equation}
\label{eqn:analysis-3}
    \frac{\partial\mathcal{E}}{\partial x_l} = \frac{\partial\mathcal{E}}{\partial x_L}\frac{\partial x_L}{\partial x_l} =
    \frac{\partial\mathcal{E}}{\partial x_L}(1 + \frac{\partial}{\partial x_l}\sum_{i=l}^{L-1} f_{RT}(x_i))
\end{equation}

Eqn.~\ref{eqn:analysis-3} indicates that the gradient $\frac{\partial\mathcal{E}}{\partial X_l}$ can be decomposed into two additive terms: a term of $\frac{\partial\mathcal{E}}{\partial X_L}$ that propagates information directly back to any shallower $l$-th block without concerning how complex $\frac{\partial}{\partial x_l}\sum_{i=l}^{L-1} f_{RT}(x_i))$ would be, and another term of $\frac{\partial\mathcal{E}}{\partial X_L}(\frac{\partial}{\partial X_l}\sum_{i=l}^{L-1} f_{RT}(X_i))$ that propagates through the Transformer blocks. 
% The additive term of $\frac{\partial\mathcal{E}}{\partial X_L}$ ensures that loss information is directly propagated back to any shallower block l. 
The equation also suggests that
% which explains 
% \minjia{It allows it to propagates to any lower block, but it does not say which one. Does it mean it propagates to all lower blocks? If that's the case, then what's the point of adding stochastic depth? What is the interplay of these two techniques? One thing seems to be clear: by adding stochastic depth, the second term becomes a stochastic sum of skipped and non-skipped layers. The answer perhaps lies in a better understanding of the stochastic depth paper.}
% iv) Eqn.~\ref{eqn:analysis-3} also suggests that 
% why identity mapping reordering avoids unbalanced gradients, because 
it is unlikely for the gradient $\frac{\partial}{\partial X_l}$ to be canceled out for a mini-batch, and in general the term $\frac{\partial}{\partial X_l}\sum_{i=l}^{L-1} f_{RT}(X_i)$ cannot be always -1 for all samples in a mini-batch.  
% Furthermore, the Transformer block with PreLN preserves the norm of the gradient in the backward path. 
This explains why the gradients of Transformer layers in Fig.~\ref{fig:stability-gradient-norm} become more balanced and do not vanish after identity mapping reordering. In contrast, the PostLN architecture has a series of layer normalization operations that constantly alter the signal that passes through the skip connection and impedes information propagation, causing both vanishing gradients and training instability.
Overall, PreLN results in several useful characteristics such as avoiding vanishing/exploding gradient, stable optimization, and performance gain. 

\section{PreLN From the View of Unrolled Iterative Refinement}
\label{sec:unrolled-analysis}

From a theoretical point of view~\cite{iterative-estimation}, a noisy estimate for a representation by the first Transformer layer should, on average, be correct even though it might have high variance. The unrolled iterative refinement view says if we treat "identity mapping" (as in PreLN) as being an unbiased estimator for the target representation, then beyond the first layer, the subsequent Transformer layer outputs $x_i^n$ (e.g., $i \in {2...L}$) are all estimators for the same latent representation $H^n$, where $H^n$ refers to the (unknown) value towards which the $n$-th representation is converging. The unbiased estimator condition can then be written as the expected difference between the estimator and the final representation: 
\begin{equation}
    \underset{x \in X}{\mathds{E}}[x_i^n - H^n] = 0
\end{equation}
% Note that both the $x_i^n$ and $H_n$ depend on the samples x of the data-generating distribution X and are thus random variables. The fact that they both depend on the same x is also the reason we need to keep them within the same expectation and cannot just write $\mathds{E}[x_i^n] = H^n$. 

With the PreLN equation, it follows that the expected difference between outputs of two consecutive layers is zero, because
\begin{equation}
    {\mathds{E}}[x_i^n - H^n] - {\mathds{E}}[x_{i-1}^n - H^n] = 0 \Rightarrow {\mathds{E}}[x_i^n - x_{i-1}^n] = 0
\end{equation}

If we write representation $x_i^n$ as a combination of $x_{i-1^n}$ and a residual ${f_{RT}}^n$, it follows from the above equation that the residual has to be zero-mean:
\begin{equation}
    x_i^n = x_{i-1}^n + {f_{RT}}^n \Rightarrow {\mathds{E}}[{f_{RT}}^n] = 0
\end{equation}
which we have empirically verified to be correct, as shown in Figure~\ref{fig:grad-norm-preserving}. Therefore, PreLN ensures that the expectation of the new estimate will be correct,  and the iterative summation of the residual functions in the remaining layers determines the variance of the new estimate ${\mathds{E}}[{F_{RT}}_i]$.

\paragraph{The effect of learning rates on downstream tasks.}
We focus on evaluating larger datasets and exclude very small datasets, as we find that the validation scores on those datasets have a large variance for different random seeds. 

For fine-tuning models on downstream tasks,
% $\in$ \{16, 32\}
% a limited hyperparameter sweep for each task, 
% and sweep learning rates $\in$ \{1e-5, 3e-5, 5e-5, \}, 
we consider training with batch size 32 and performing a
linear warmup for the first 10\% of steps followed by
a linear decay to 0. We fine-tune for 5 epochs and
perform the evaluation on the development set. 
% We apply The rest of the hyperparameters remain the same as during pre-training.
% In this setting, 
We report the median development
set results for each task over five random initializations, without model ensemble.

Results are visualized in Fig.~\ref{fig:fine-tune-heatmap}, which shows that the baseline is less robust on the choice of learning rates. Specifically, the fine-tuning results are often much worse with a large learning rate. In comparison, \pst is more robust and often achieves better results with large learning rates. 
% to ensure that we do not unlearn the previously acquired knowledge.

% \begin{figure}[ht!]
% \centering
% \small
% \subfloat[MNLI-m]{\includegraphics[scale=0.33, keepaspectratio=true]{figs/heatmap/heatmap-MNLI-m}\label{fig:heatmap-MNLI-m}}
% \subfloat[MNLI-mm]{\includegraphics[scale=0.33, keepaspectratio=true]{figs/heatmap/heatmap-MNLI-mm}\label{fig:heatmap/heatmap-MNLI-mm}}
% \subfloat[QNLI]{\includegraphics[scale=0.33, keepaspectratio=true]{figs/heatmap/heatmap-QNLI}\label{fig:heatmap/heatmap-QNLI}} \\
% \subfloat[QQP]{\includegraphics[scale=0.33, keepaspectratio=true]{figs/heatmap/heatmap-QQP}\label{fig:heatmap/heatmap-QQP}} 
% \subfloat[RTE]{\includegraphics[scale=0.33, keepaspectratio=true]{figs/heatmap/heatmap-RTE}\label{fig:heatmap/heatmap-RTE}}
% \subfloat[SST-2]{\includegraphics[scale=0.33, keepaspectratio=true]{figs/heatmap/heatmap-SST-2}\label{fig:heatmap/heatmap-SST-2}} 
% \caption{The fine-tuning results at different checkpoints.}
% \label{fig:fine-tune-heatmap}
% \end{figure}

% \section{Additional Results}
% \label{sec:extra}

% \begin{figure}
%     \centering
%     \includegraphics[scale=0.7]{figs/varying-keep-ratio.png}
%     \caption{Convergence curves varying the keep ratio $\bar{\theta}$.}
%     \label{fig:varying-keep-ratio}
% \end{figure}

\end{document}